\documentclass[conference]{IEEEtran}
\IEEEoverridecommandlockouts
\usepackage{amsmath,amssymb,amsfonts}
\usepackage{graphicx}
\usepackage{textcomp}
\usepackage{xcolor}
\usepackage{booktabs} 
\usepackage[linesnumbered,ruled,vlined]{algorithm2e}
\usepackage{booktabs}

\usepackage{url}
\usepackage{physics}  
\usepackage{subfigure}
\usepackage{cite}
\usepackage{multirow}
\usepackage{bm}
\usepackage{comment}
\usepackage{braket}
\usepackage{mathtools}
\usepackage{enumerate}
\usepackage{enumitem}
\usepackage{empheq}
\usepackage{calc}
\usepackage{balance}
\usepackage{dsfont}
\usepackage{color, colortbl}
\usepackage[first=0,last=9]{lcg}
\graphicspath{{images/}}

\def\BibTeX{{\rm B\kern-.05em{\sc i\kern-.025em b}\kern-.08em
    T\kern-.1667em\lower.7ex\hbox{E}\kern-.125emX}}
\begin{document}

\title{Heterogeneous Stream-reservoir Graph Networks with Data Assimilation} 

\author{Shengyu Chen$^1$, Alison Appling$^2$, Samantha Oliver$^2$, Hayley Corson-Dosch$^2$, Jordan Read$^2$,\\ Jeffrey Sadler$^2$, Jacob Zwart$^2$, Xiaowei Jia$^1$\\
$^1$University of Pittsburgh, $^2$U.S. Geological Survey
}

\maketitle

\begin{abstract}
Accurate prediction of water temperature in streams is critical for monitoring and understanding biogeochemical and ecological processes in streams. Stream temperature is affected by weather patterns (such as solar radiation) and water flowing through the stream network. Additionally, stream temperature can be substantially affected by water releases from man-made reservoirs to downstream segments. In this paper, we propose a heterogeneous recurrent graph model to represent these interacting processes that underlie stream-reservoir networks and improve the prediction of water temperature in all river segments within a network. Because reservoir release data may be unavailable for certain reservoirs, we further develop a data assimilation mechanism to adjust the deep learning model states to correct for the prediction bias caused by reservoir releases. A well-trained temporal modeling component is needed in order to use adjusted states to improve future predictions. Hence, we also introduce a simulation-based pre-training  strategy to enhance the model training. Our evaluation for the Delaware River Basin has demonstrated the superiority of our proposed method over multiple existing methods. We have extensively studied the effect of the data assimilation mechanism under different scenarios. Moreover, we show that the proposed method using the pre-training strategy can still produce good predictions even with limited training data.  
\end{abstract}


%


\section{Introduction}

A healthy aquatic ecosystem is key to the sustainability of our planet. These ecosystems are under growing pressures caused by local demand to global food, water, and energy networks. 
Water temperature in streams is a “master variable”~\cite{brett1971energetic} for many important aquatic outcomes, including the suitability of aquatic habitats, evaporation rates, and greenhouse gas exchange. Studying the above processes benefits from the capacity to automatically predict and monitor stream temperature over multiple river segments in a stream network and under a frequent time interval. Accurate predictions can also help with the assessment of dependencies between streams and human infrastructures such as dams and reservoirs while also providing valuable information to aid in decision making for resource managers. 
%

Water temperature in streams 
is affected by a combination of processes including weather (e.g., air temperature, solar radiation, precipitation), interactions between connected river segments in the stream network, and the process of water release from reservoirs. 
In particular, multiple river segments in a stream network often exhibit different thermodynamic patterns driven by differences in catchment characteristics (e.g. slope, soil characteristics, groundwater influence) and climate drivers (e.g. air temperature, precipitation). These segments are also affected by  the water advected from upstream river segments and reservoirs.

Existing works on this prediction problem are mostly focused on building physics-based models~\cite{theurer1984instream,markstrom2015prms} and data-driven models~\cite{moshe2020hydronets,jia2021physics} to simulate how stream temperature changes given variations in climate drivers (e.g., solar radiation, precipitation, and air temperature) and catchment characteristics, and due to the influence of the river network topology.  Recently, researchers have also started to further explore the stream network structure and leverage the advances in graph neural networks to capture such spatial dependencies~\cite{jia2021physics,moshe2020hydronets}. However, these methods are not designed for modeling the dependencies between streams and reservoirs. Reservoirs are man-made lakes formed upstream from  dams. Reservoir managers determine how much water to release into the downstream river network. The release of cold water from the bottom of the reservoirs 
then reduces downstream water temperatures. 
Existing approaches do not capture interactions between reservoirs and streams and thus have degraded performance in predicting water temperature for river networks with reservoirs. 


In this work, we develop a new data-driven method to model stream-reservoir networks. There are several challenges faced by existing machine learning algorithms when applied to this problem. First, although they are connected to rivers, the physical properties of reservoirs are very different than those of river segments. For example, reservoirs commonly have stratified layers with different temperature  while streams are  shallower and usually assumed to be well-mixed. Stream water flowing into a reservoir affects the reservoir's temperature, and water release from reservoirs also greatly affects the temperature of downstream river segments. Second, given the substantial personnel and material cost in collecting water temperature observations, there are often only a handful of river segments in a network that are monitored; thus, there are limited labeled data to train ML models. Third, reservoir data, e.g., when and how much water is released, are often unavailable for certain reservoirs due to privacy and security issues. This brings stochasticity  to 
the relation between input climate drivers and observed water temperature, and makes it difficult for data-driven methods to capture such relation. 

To address these challenges, we propose a model termed Heterogeneous Recurrent Graph Networks (HRGN). This model aims to capture the distinct behavior of river segments and reservoirs, while also simultaneously learning the temporal thermodynamics and interactions amongst streams and reservoirs. We also introduce a data assimilation mechanism to efficiently adjust the model state in order to correct the predictive bias caused by missing reservoir release data. The state adjustment is implemented using an invertible network structure and can be trained together with the HRGN model. Because real-world observations may be limited for training advanced deep learning models, 
we also discuss a strategy to pre-train a subset of model components using physical simulations. The pre-trained model then requires much less data to fine-tune itself to a quality model.

We evaluate the proposed method in the Delaware River Basin. The results demonstrate the superiority of the proposed method in predicting water temperature when reservoirs are present in the stream network. Also, we show the effectiveness of the state adjustment mechanism when the water release information is not available.  We have extensively studied the adjustment of the model state using different amounts of data and different time delay. Finally, we show the effectiveness of pre-training when we are not provided with sufficient observations for training.

\section{Related Work}

Our proposed method has multiple components, including the graph neural networks for modeling interacting processes, model state adjustment (or data assimilation), and pre-training using physical simulations. These components have been extensively studied under different contexts. Here we provide a brief review for each of these components.

\subsection{Graph Neural Networks for Modeling Interacting Processes}
Graph neural networks, e.g., Graph Convolutional Networks (GCN)~\cite{kipf2016semi} and  GraphSAGE~\cite{hamilton2017inductive}, have found immense success in commercial applications due to their power in automatically representing information propagation amongst multiple objects. Recently, graph neural networks have also been applied to multiple scientific problems and shown improved predictive performance~\cite{qi2019hybrid,xie2018crystal,zhu2020understanding}. These advances show promise for modeling interacting processes in complex physical systems, which commonly requires substantial efforts in calibration in traditional physics-based modeling approaches. 
Graph neural networks have also been applied to modeling water temperature and streamflow in river networks~\cite{jia2021physics,moshe2020hydronets}. Although they have shown improved performance compared to existing physics-based models, they are mostly evaluated in stream regions without reservoirs. The performance of these methods can be severely affected when reservoirs are present in the stream networks but unaccounted for in the graph network.   


A heterogeneous graph is commonly used to represent multiple types of connections amongst multiple types of instances~\cite{shi2016survey}. Neural network models have also been developed to represent such a graph structure and discover knowledge from heterogeneous data~\cite{zhang2019heterogeneous,wang2019heterogeneous,zhu2020hgcn}. However, this powerful model has not been used to represent multiple complex interactions amongst different types of processes in scientific problems. The nature of scientific studies also benefits from adaptation of these neural network models based on scientific knowledge to better represent the influence amongst processes.


\subsection{Data Assimilation}
Data assimilation has been widely used in physics-based modeling approaches with the aim to optimally adjust the model state of a system with recent observations. There are many data assimilation approaches used for physics-based models and they differ in their assumptions about the distributions of the model and data predictions and speed of computation. The Kalman filter, which iteratively  estimates the next state through a forward process and then updates the state using new observations,  
is a popular data assimilation technique for physics-based models. Many variations of the Kalman filter have also been implemented~\cite{evensen2009data}. Despite their extensive use, these approaches can be computationally expensive to implement when we have a large state space and/or non-linear system dynamics. Recently, researchers have started to use neural networks as an alternative way for implementing data assimilation~\cite{fang2020near,brajard2020combining}. Note that this is different from online incremental learning in that model parameters remain the same but only states are changed. The intuition of data assimilation is to adjust the model state when they are disturbed by external factors that are not captured by input features. For example, Brajard et al.~\cite{brajard2020combining} propose a two-stage process: (1) in the training phase, model parameters (i.e., network weights) are trained using observations, and (2) in the data assimilation phase, model parameters are fixed and the model state is optimized to match observations using back-propagation. 

\subsection{Pre-training Using Physical Simulations}
Neural network models require an initial choice of model parameters before training. Poor initialization can cause models to anchor in local minima, which is especially true for complex models. Transfer learning has been widely used in computer vision~\cite{tan2018survey}, where the pre-trained models from a related large-scale dataset are fine-tuned with limited training data to fit the target task. 
In scientific problems, we can use a physics-based model's simulated data to pre-train the ML model.
This approach can also alleviate data paucity issues.
Read et al.~\cite{read2019process} show that recurrent neural networks after being pre-trained using simulation outputs are able to generalize better to unseen scenarios than pure physics-based models or machine learning models.  
Jia et al. also extensively discuss this strategy~\cite{jia2021physics_tds}. They pre-train their Physics-Guided Recurrent Neural Network (PGRNN) models for lake temperature modeling on simulated data generated from a physics-based model and fine-tune it with little observed data. They show that pre-training 
can 
significantly reduce the training data needed for a quality model. Moreover, they show that machine learning models can still benefit from such pre-training process by learning general, physically consistent patterns (e.g., seasonal cycles) even when physical simulations are biased. 


\section{Method}

In this section, we first introduce the proposed heterogeneous recurrent graph model architecture to represent stream-reservoir networks. Then we introduce the data assimilation strategy to improve the prediction when reservoir release data are missing. Finally, we discuss a pre-training method using physical simulations in order to improve the performance of prediction and data assimilation when we have limited training data samples.

\begin{figure} [!h] 
\hspace{-0.2in}
\subfigure[]{ \label{fig:a}{}
\includegraphics[width=0.32\columnwidth]{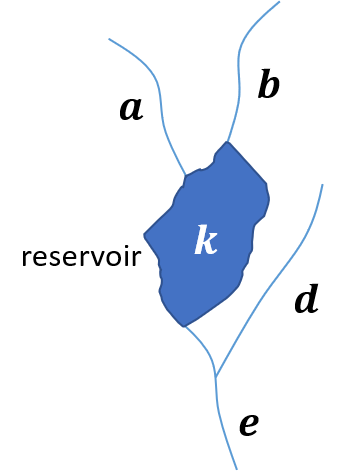}
}\hspace{-0.1in}
\subfigure[]{ \label{fig:b}{}
\includegraphics[width=0.32\columnwidth]{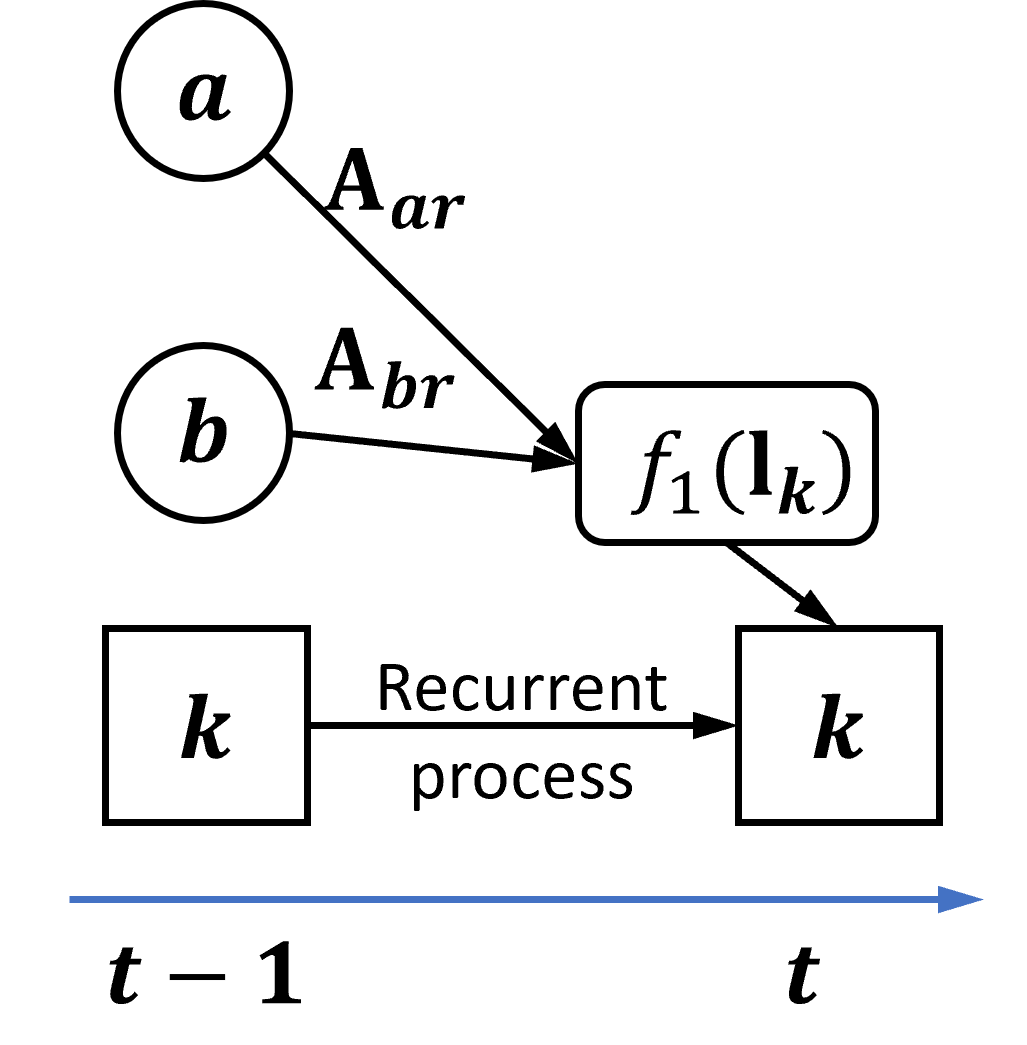}
}\hspace{-.1in}
\subfigure[]{ \label{fig:b}{}
\includegraphics[width=0.32\columnwidth]{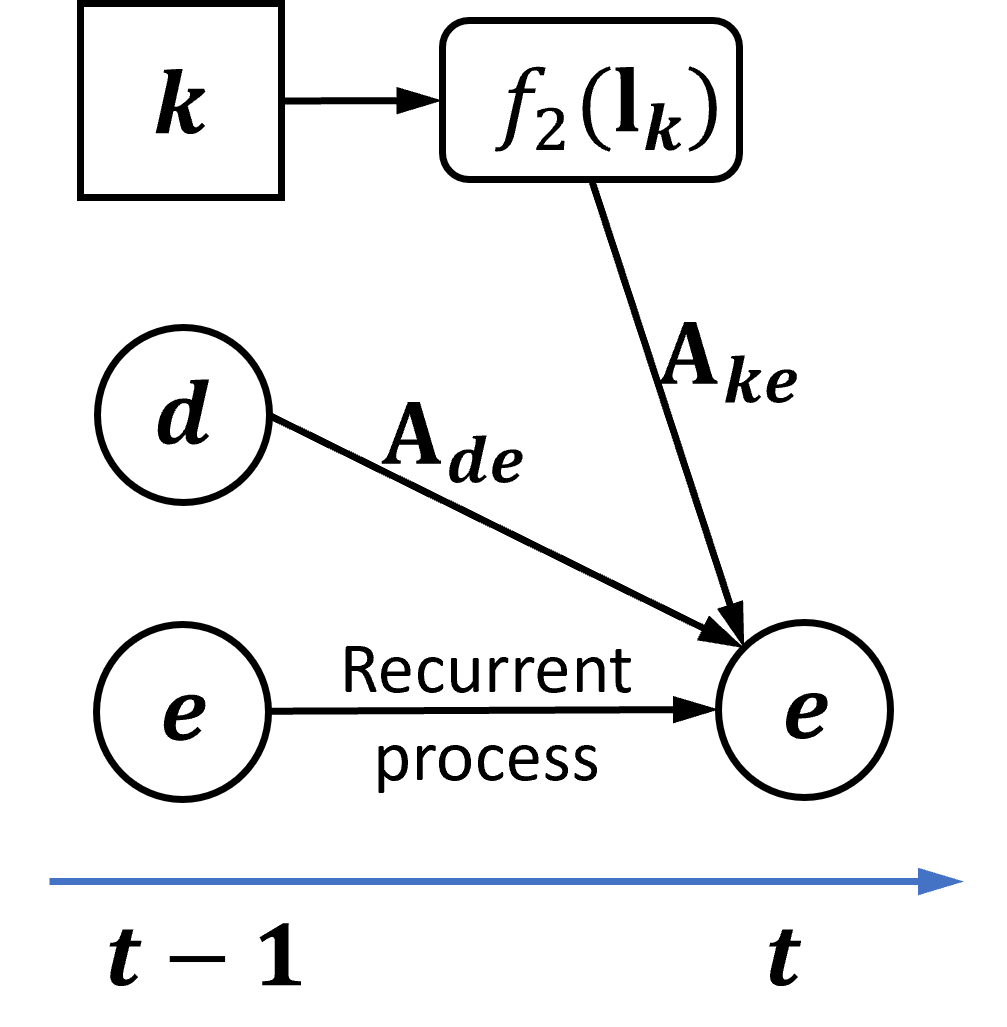}
}
\caption{Overall structure of the HRGN model. (a) An example stream-reservoir network. (b) The recurrent modeling process for the reservoir node $k$. (c) The recurrent modeling process for the river segment node $e$. Here $\textbf{l}_k$ represents the characteristics of the reservoir $k$ and $\textbf{A}$ is the adjacency matrix.}
\label{fig:hrgn}
\end{figure}

\subsection{Heterogeneous Recurrent Graph Networks (HRGN)}

A stream-reservoir network is a dynamical system which consists of multiple river segments and reservoirs. The reservoirs are very different from rivers in that they commonly have a stratified temperature profile over a deeper structure while rivers are commonly much shallower and assumed to be well-mixed. 
Stream temperature is affected by weather (e.g., change of air temperature and precipitation) as well as water flowing from upstream river segments and reservoirs and from groundwater. Similarly, reservoir inflows from upstream river segments can also change the temperature profile of the reservoir.  
We build a heterogeneous recurrent graph architecture to represent dependencies amongst river segments, reservoirs, and climate drivers while also capturing different natures of streams and reservoirs. 
Our objective is to extract spatial and temporal contextual information using this graph structure and improve the prediction of water temperature over multiple river segments in the network.

In particular, we consider $N$ river segments and $M$ reservoirs in a river network. For each river segment $i$, we are provided with input features over multiple time steps $\textbf{X}_i=\{\textbf{x}_{i}^{1}, \textbf{x}_{i}^2, ..., \textbf{x}_{i}^T\}$.  Here input features $\textbf{x}_{i}^{t}$ are a $D$-dimensional vector, which includes climate drivers and geometric parameters of the segment (more details can be found in Section~\ref{sec:dataset}). For each reservoir $k$, We are  provided with its static meta-features $\textbf{l}_k$ such as the height and width of the dam. We also assume that we have access to the water release information for the reservoir $k$ at a daily scale $\{\textbf{r}_k^1,...,\textbf{r}_k^T\}$, where each $\textbf{r}_k^t$ represents the amount of water released under different release types (conservation release, direct release, and spill release) at each time step $t$.


A heterogeneous graph refers to a graph structure that contains multiple types of nodes and edges. In this problem, we build graph structure $\mathcal{G} = \{\mathcal{V},\mathcal{E},\textbf{A}\}$. Here the node set $\mathcal{V}=\{\mathcal{V}_s,\mathcal{V}_r\}$ contains the set of river segments $\mathcal{V}_s$ and reservoirs $\mathcal{V}_r$. The edge set $\mathcal{E}=\{\mathcal{E}_{ss},\mathcal{E}_{sr},\mathcal{E}_{rs}\}$ contains three types of edges among river segments and reservoirs. Specifically, $\mathcal{E}_{ss}$ represents the edges between pairs of segments $(i,j)$ where the segment $i$ is anywhere upstream from the segment $j$, $\mathcal{E}_{sr}$ represents the edges between river segments and their downstream reservoirs, and $\mathcal{E}_{rs}$ represents the edges between reservoirs and their downstream river segments. In the following discussion, we use $\mathcal{N}(i)$ and $\mathcal{M}(i)$ to represent river segments and reservoirs, respectively, that are upstream from the river segment $i$, i.e., $j\in \mathcal{N}(i)$ if $(j,i)\in \mathcal{E}_{ss}$ and $k\in \mathcal{M}(i)$ if $(k,i)\in \mathcal{E}_{rs}$. We also use $\mathcal{S}(k)$ to represent river segments upstream from the reservoir $k$, i.e., any river segment $i$ such that $(i,k)\in \mathcal{E}_{sr}$. 
The matrix $\textbf{A}$ represents the adjacency level between each pair of river segments or between river segments and reservoirs in the graph. Specifically, $\textbf{A}_{ij}=0$ means there is no connection from node $i$ to node $j$ and a higher value of $\textbf{A}_{ij}$ indicates that the node $i$ is closer to node $j$ in terms of the stream distance. 
More details of how we generate the adjacency matrix are discussed in Section~\ref{sec:dataset}.


We now describe the recurrent process of generating the hidden representation $\textbf{h}_i^t$ from $\textbf{h}_i^{t-1}$ for each river segment $i$. We use the hidden representation to encode spatial and temporal contextual information and generate prediction of water temperature at time $t$. We repeat this process for multiple time steps over the entire sequence. 
For each river segment $i$, its water temperature at time $t$ depends on the information from four sources: (1) the current climate inputs, (2) the previous state for the segment $i$, (3) its upstream river segments, and (4) its upstream reservoirs. When modeling the stream temperature dynamics, we also need to consider the water dynamics in reservoirs. Hence, we maintain a series of states for each river segment ($\textbf{c}_i^t$) and reservoir ($\textbf{cr}_k^t$).

Because water flow from upstream river segments can change the temperature of reservoirs, we update the reservoir state $\textbf{cr}_k^{t}$  by incorporating the influence from  its upstream river segments at the previous time step $t-1$. Moreover, the change of reservoir temperature given such influence also  depends on the characteristics of the reservoir (e.g., the geometry of reservoirs). Hence, we use the static features to filter the influence from its upstream river segments (represented as $\mathcal{S}(k)$) before updating the reservoir state. This process can be described as follows:
\begin{equation}
\textbf{cr}_{k}^{t} = \text{tanh}(\textbf{W}_{cr} \textbf{cr}_{k}^{t-1}+ f_1(\textbf{l}_k)\otimes \sum_{i\in \mathcal{S}(k)}\textbf{A}_{ik}\textbf{h}_i^{t-1} +\textbf{b}_{cr}),
\label{eq:res-state}
\end{equation}
where $\textbf{W}_{cr}$ and $\textbf{b}_{cr}$ are model parameters, $\otimes$ represents the element-wise product, the function $f_1(\cdot)$ transforms the static meta-features of the reservoir to the same dimension with hidden variables, which is then used to filter the influence from upstream river segments through the element-wise product. We implement $f_1(\cdot)$ using fully connected layers. Here the influence of each upstream river segment is also weighted by its adjacency level to the reservoir.


The water temperature of each river segment $i$ can be affected by the water release from upstream reservoirs. At time $t-1$, the model extracts latent variables, which embed the influence of upstream reservoirs to the downstream river segment $i$. We refer to these latent variables as transferred variables.
The influence of reservoir release to downstream segments is determined by  the amount of water release, the temperature of released water, and the reservoir characteristics (e.g., dam height and length).
Hence, for the river segment $i$, we compute the transferred variables from its upstream reservoirs (represented as $\mathcal{M}(i)$), as follows:
\begin{equation}
\textbf{p}_{i}^{t-1} = \text{tanh}(\textbf{W}_p\!\!\sum_{k \in \mathcal{M}(i)}\!\!\textbf{A}_{ki}f_2(\textbf{l}_k)\otimes(\textbf{W}_{p}^r\textbf{r}_k^{t-1}+\textbf{W}_p^{cr} \textbf{cr}^{t-1}_{k})+\textbf{b}_p),
\end{equation}
where \{$\textbf{W}_p$,$\textbf{W}_{p}^r$, $\textbf{W}_p^{cr}$, $\textbf{b}_p$\} are model parameters,  $f_2(\cdot)$ is also used to convert static features of the reservoir to the filtering variables and is implemented using fully connected layers. 


We now consider the influence from upstream river segments to downstream segments.   Specifically, for each river segment $i$, we compute the transferred variables from its upstream river segments (represented as $\mathcal{N}(i)$)  as follows:
\begin{equation}
\textbf{q}_{i}^{t-1} = \text{tanh}(\textbf{W}_q\sum_{j \in \mathcal{N}(i)}\textbf{A}_{ji}\textbf{h}_{j}^{t-1}+\textbf{b}_q), 
\end{equation}
where $\textbf{W}_q$ and $\textbf{b}_q$ are mode parameters.

Similar to Long-Short Term Memory (LSTM), we also generate a candidate cell state $\bar{\textbf{c}}_i^t$ by combining climate drivers at the current time step $\textbf{\textbf{x}}_i^t$ and the hidden representation at previous time step $\textbf{\textbf{h}}_i^{t-1}$, as follows: 
\begin{equation}
\begin{aligned}
\bar{\textbf{c}}_i^t &= \text{tanh}(\textbf{W}_c^h \textbf{h}_i^{t-1} + \textbf{W}_c^x \textbf{x}_i^t+\textbf{b}_c),
\end{aligned}
\end{equation}
where \{$\textbf{W}_c^h$,$\textbf{W}_c^x$,$\textbf{b}_c$\} are model parameters.

Then we generate four sets of  gating variables, forget gating variables $\textbf{gf}^t_i$, input gating variables $\textbf{gi}^t_i$, upstream reservoir gating variables $\textbf{gr}^t_i$, and upstream  river segment gating variables $\textbf{gs}^t_i$. These gating variables are used to filter the information passed from the previous time step, the current time step, upstream reservoirs, and upstream river segments, respectively. Formally, these gating variables are computed using the sigmoid function $\sigma(\cdot)$ as follows: 
\begin{equation}
\begin{aligned}
\textbf{gf}_i^t &= \sigma(\textbf{W}_f^h \textbf{h}_i^{t-1} + \textbf{W}_f^x \textbf{x}_i^t+\textbf{b}_f),\\
\textbf{gi}_i^t &= \sigma(\textbf{W}_g^h \textbf{h}_i^{t-1} + \textbf{W}_g^x \textbf{x}_i^t+\textbf{b}_g),\\
\textbf{gr}_i^t &= \sigma(\textbf{W}_r^p \textbf{p}^{t-1}_{i} + \textbf{W}_r^x \textbf{x}_i^t+\textbf{b}_r),\\
\textbf{gs}_i^t &= \sigma(\textbf{W}_s^q \textbf{q}^{t-1}_{i} + \textbf{W}_s^x \textbf{x}_i^t+\textbf{b}_s),
\end{aligned}
\end{equation}
where \{$\textbf{W}_f^h$, $\textbf{W}_f^x$, $\textbf{W}_g^h$, $\textbf{W}_g^x$, $\textbf{W}_r^p$, $\textbf{W}_r^x$, $\textbf{W}_s^q$, $\textbf{W}_s^x$, ,$\textbf{b}_f$, $\textbf{b}_g$, $\textbf{b}_r$,$\textbf{b}_s$\} are model parameters.


Once we obtain the gating variables, we can use them to filter the information from the previous time ($\textbf{c}_i^{t-1}$), the current time step ($\bar{\textbf{c}}_i^t$) and the transferred variables from upstream reservoirs and river segments via element-wise product, and combine the filtered information to compute the model state at time $t$. This process can be expressed as follows:  
\begin{equation}
\begin{aligned}
\textbf{c}_i^t &= \text{tanh}(\textbf{gf}_i^t\otimes \textbf{c}_i^{t-1}+\textbf{gi}_i^t\otimes\bar{\textbf{c}}_i^t+\textbf{gr}_i^t\otimes \textbf{p}^{t-1}_i+\textbf{gs}_i^t\otimes \textbf{q}^{t-1}_i).
\end{aligned}
\label{eq:hid}
\end{equation}

According to this equation, the change of the model state given the inputs over space and time is conditioned on the gating variables $\textbf{gf}_i^t$, $\textbf{gi}_i^t$,  $\textbf{gr}_i^t$, and $\textbf{gs}_i^t$.  This is analogous to the evolution of a dynamical system, in which state variables change over time in response to influences from different sources filtered by specific physical conditions.  

After obtaining the model state $\textbf{c}_i^t$, we generate the output gating variables $\textbf{o}^t_i$ and use them to filter the model state to generate the hidden representation $\textbf{h}^t$, as follows:
\begin{equation}
\begin{aligned}
\textbf{o}_i^t &= \sigma(\textbf{W}_o^h \textbf{h}_i^{t-1} + \textbf{W}_o^x \textbf{x}_i^t+\textbf{b}_o),\\
\textbf{h}_i^t &= \textbf{o}_i^t\otimes (\textbf{c}_i^t),
\end{aligned}
\label{eq:hidden}
\end{equation}
where \{$\textbf{W}_o^h$,$\textbf{W}_o^x$,$\textbf{b}_o$\} are model parameters. 


Finally, we generate predicted target variables $\hat{\textbf{y}}_i^t$ from the hidden representation, as follows:
\begin{equation}
\hat{\textbf{y}}_i^t = \textbf{W}_y\textbf{h}_i^t+\textbf{b}_y,
\label{eq:predict}
\end{equation}
where $\textbf{W}_y$ and $\textbf{b}_y$ are model parameters. 

The loss function of HRGN is defined  as the mean squared loss between true observations 
$\textbf{Y}=\{\textbf{y}_{i}^{t}\}$ and predicted values. The loss is only measured at certain time steps and locations for which observations are available, as follows:
\begin{equation}
    \mathcal{L}_{\text{HRGN}}(\hat{\textbf{Y}},\textbf{Y}) = \frac{1}{|\textbf{Y}|} \sum_{\{(i,t)|\textbf{y}_{i}^{t}\in \textbf{Y}\}} (\textbf{y}_{i}^{t}-\hat{\textbf{y}}_{i}^{t})^2.
    \label{eq:std_loss}
\end{equation}


\subsection{Data Assimilation Mechanism}

Reservoir water release is controlled by independent water management agencies and such information is often not available due to the privacy and security reasons. The limited water release data may directly degrade the performance of the proposed model. 
For example, if water managers release large volume of cold water at the beginning of summer, the water temperature for downstream segments can become much colder in the next few days than they would have been without the water release. Hence, the model may overestimate water temperature if it is not aware of such water release information.

\begin{figure} [!t] 
\centering
\includegraphics[width=1\columnwidth]{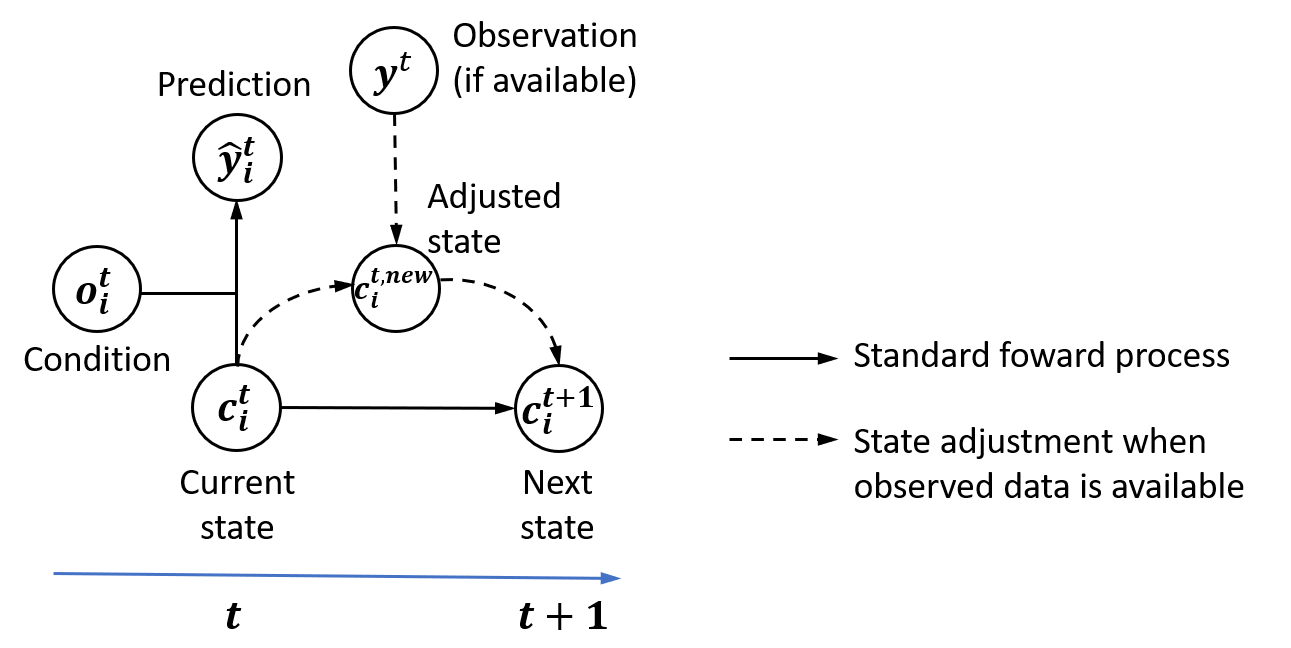}
\caption{The process of data assimilation for adjusting the model state. The adjusted state at time $t$ will not change the predictions at $t$ but is only used to estimate the next state $\textbf{c}_i^{t+1}$ with the hope that the model can get better performance. }
\label{fig:model_adjust}
\end{figure}

To overcome this limitation, we propose a new data assimilation strategy that aims to adjust the model state if we have downstream water temperature observations during the prediction phase. Here we assume that the water release will cause bias in model predictions. We aim to leverage available observations (collected after making predictions) to correct the model bias. This process is shown in Fig.~\ref{fig:model_adjust}. Specifically, after we predict temperature at time $t$ for all the river segments, if we have temperature observations at $t$ for the  segment $i$, we can use  these observations to adjust the model state $\textbf{c}_i^t$ before it is used to compute the  state $\textbf{c}_i^{t+1}$ and prediction $\hat{\textbf{y}}_i^{t+1}$ at the next time step. 

In practice, we may need to adjust the model state after a few time steps, which allows sufficient time for reviewing and cleaning collected observations. Hence, we implement two versions of model adjustment in our experiments. The first version is to update the model state immediately after we collect observations. In the second version,  an update period is specified (e.g., 7 days) and the state update is conducted only by the end of each update period.  
In the following, we will describe the method used for immediate update. The method presented herein can be easily adapted to the delayed update using an update period. 

According to Eqs.~\ref{eq:hidden}-\ref{eq:predict}, the predictions $\hat{\textbf{y}}_i^t$ are computed as a function of the model state $\textbf{c}_i^t$ while also being conditioned on the output gating variables $\textbf{o}_i^t$, as follows:
\begin{equation}
\hat{\textbf{y}}_i^t = f(\textbf{c}_i^t|\textbf{o}_i^t),
\label{eq:general_prd}
\end{equation}

We aim to build a reverse structure that can update the model state given a new observations $\textbf{y}_i^t$ (obtained after we make predictions), i.e., $\textbf{c}_i^{t,new} = f^{-1}(\textbf{y}_i^t|\textbf{o}_i^t)$.
We first build an invertible structure to map the model state $\textbf{c}_i^t$ to the hidden representation $\textbf{h}_i^t$. Here we  use the invertible network layers introduced in previous literature~\cite{dinh2016density,ardizzone2019guided} to replace the  mapping  defined in Eq.~\ref{eq:hidden}. 
In particular, we implement the invertible layers by equally dividing the state variables into two parts, as $\textbf{c}_{i,1}^t$ and $\textbf{c}_{i,2}^t$. Then they are converted to the hidden representation $\textbf{h}_i^t$ as follows:
\begin{equation}
\begin{aligned}
    \textbf{h}_{i}^t &= [\textbf{h}_{i,1}^t,\textbf{h}_{i,2}^t],\\
    \textbf{h}_{i,1}^t &= \textbf{c}_{i,1}^t \otimes \text{exp}(s_1([\textbf{c}_{i,2}^t,\textbf{o}_i^t])) + g_1([\textbf{c}_{i,2}^t,\textbf{o}_i^t])  \\
    \textbf{h}_{i,2}^t &= \textbf{c}_{i,2}^t \otimes \text{exp}(s_2([\textbf{h}_{i,1}^t,\textbf{o}_i^t])) + g_2([\textbf{h}_{i,1}^t,\textbf{o}_i^t]),  \\
\end{aligned}
\label{eq:foward}
\end{equation}
where $s_1$,$s_2$,$g_1$,$g_2$ can be arbitrary functions and we implement them using fully connected layers. The process of splitting $\textbf{h}_{i}^t$ and  $\textbf{c}_{i,1}^t$ is required to ensure the invertiable property of the network.  Given a new hidden representation $\textbf{h}_i^t$, we can estimate the updated model state using an inverted process:
\begin{equation}
\begin{aligned}
    \textbf{c}_{i}^{t,new} &= [\textbf{c}_{i,1}^{t,new},\textbf{c}_{i,2}^{t,new}],\\
    \textbf{c}_{i,2}^{t,new} &= (\textbf{h}_{i,2}^t - g_2([\textbf{h}_{i,1}^t,\textbf{o}_i^t])) \oslash \text{exp}(s_2([\textbf{h}_{i,1}^t,\textbf{o}_i^t])),  \\
    \textbf{c}_{i,1}^{t,new} &= (\textbf{h}_{i,1}^t - g_1([\textbf{c}_{i,2}^{t,new},\textbf{o}_i^t])) \oslash \text{exp}(s_1([\textbf{c}_{i,2}^{t,new},\textbf{o}_i^t])).  \\
\end{aligned}
\label{eq:invert}
\end{equation}

It can be seen that the inverted process from $\textbf{h}_i^t$ to $\textbf{c}_i^t$ is done without additional parameters but the functions $s_1$,$s_2$,$g_1$,$g_2$ learned from the forward process. Moreover, one can stack multiple invertible layers by repeating the forward process (Eq.~\ref{eq:foward}) multiple times and the mapping from $\textbf{c}_i^t$ to $\textbf{h}_i^t$ is still invertible (by multiple iteration of Eq.~\ref{eq:invert}). It is noteworthy that the invertiable layer requires $\textbf{c}_i^t$ and $\textbf{h}_i^t$ to have the same dimensionality.

We then build another fully connected layer to convert the hidden representation to the predicted outputs (same with Eq.~\ref{eq:predict}). To make this process also invertible, we create a reverted mapping from observations to the hidden representation, as $\textbf{h}_i^t = u({\textbf{y}}_i^t)$ where $u(\cdot)$ is implemented by a fully connected layer. We also include a reconstruction loss to ensure that the function $u(\cdot)$  indeed maps observations to corresponding points in the hidden space. Here we represent the forward prediction process defined by 
Eq.~\ref{eq:predict} as functions $v(\cdot)$. 
Because observations are sparse, we can use model predictions to define the reconstruction loss, which is measured as the mean squared loss (MSE) between the predictions $\hat{\textbf{y}}$ and reconstructed predictions, as follows:
\begin{equation}
    \mathcal{L}_\text{recon}=\sum_i\sum_t \text{MSE} (\hat{\textbf{y}}_i^t,v\circ u(\hat{\textbf{y}}_i^t)))/NT,
\end{equation}
where $v\circ u$ denotes the function composition. Here the reconstruction loss is only applied to the mapping from $\textbf{h}_i^t$ to $\hat{\textbf{y}}_i^t$ while the mapping from $\textbf{c}_i^t$ to $\textbf{h}_i^t$ (Eqs.~\ref{eq:foward} and \ref{eq:invert}) is automatically invertible. This helps reduce the model complexity in maintaining the invertible property and increase the chance for the model to be better generalizable. 

Then the final loss function is defined as follows:
\begin{equation}
    \mathcal{L} = \mathcal{L}_\text{HRGN}+\lambda\mathcal{L}_\text{recon},
    \label{eq:full_loss}
\end{equation}
where $\lambda$ is a hyper-parameter to balance two loss terms.

In the data assimilation process, we use observations to first estimate the hidden representation through the function $u(\cdot)$ and then compute the updated state by Eq.~\ref{eq:invert}. 
When we implement this method, we conduct training in two stages. In the first stage, we train the HRGN model by minimizing the loss function (Eq.~\ref{eq:full_loss}) without using the data assimilation mechanism. This ensures learning a good forward model. In the second stage, we use training observations (when and where available) to adjust the model state (using the function $u(\cdot)$ and Eq.~\ref{eq:invert}) after it makes predictions at each time step. The updated state is used compute the model state and predictions in the next step. Then we collect these predicted values (using updated previous states) and optimize the model using the same loss function. This training process helps adapt the model to the data assimilation scenario. When we apply the trained model to the testing period, it is used in a similar way to the second training stage, i.e., it makes predictions at each time $t$ and then gets an updated state using observations if available. It is noteworthy that we only use observations from a historical view, i.e., after we make predictions, so this strategy can be used in practice for generating future predictions. 

Compared to existing neural networks-based data assimilation method, we build an invertible structure for state adjustment and this structure can be trained together with the predictive modeling. The invertible structure can more efficiently reconstruct the model state given observations because it does not require an optimization process during the adjustment. Moreover, it can potentially produce better adjustment due to its invertible nature and fewer assumptions of data distributions and system dynamics than other data assimilation techniques.

\subsection{Model Pre-training}


The intuition of data assimilation mechanism is to adjust the model state to better estimate the state and the predictions in the future steps. This requires a well-trained model component to capture the transition of states over time (i.e., Eqs.~\ref{eq:res-state}-\ref{eq:hid}), otherwise the adjustment at previous time will not be reflected to the next state. The transition process involves a large number of parameters and requires sufficient data for effective training. However, collecting observations is often expensive in scientific applications. For example, collecting water temperature data commonly requires highly trained scientists to travel to sampling locations and deploy sensors within a stream, leading to substantial personnel and equipment costs.

To address these issues, we propose to pre-train the model using simulated data produced by a physics-based model, PRMS-SNTemp~\cite{theurer1984instream,markstrom2015prms}. PRMS-SNTemp simulates  daily stream water temperature for each river segment by solving an energy/mass balance model that accounts for the effect of inflows (upstream, groundwater, surface runoff), outflows, and surface heating and cooling on heat transfer in each river segment.

The simulated water temperature data are imperfect due to the approximations used in representing physical processes and the uncertainty in selecting model parameters, but they provide a synthetic realization of physical responses of a river network to a given set of climate drivers and flow processes. Hence, we hypothesize that the model pre-trained using sufficient simulated data can get closer to the optimal solution and requires fewer observations for fine-tuning. 

One limitation of PRMS-SNTemp is that it is purely based on physics about temperature dynamics but does not simulate the effects of human infrastructures such as reservoirs. Hence, in pre-training we cannot consider the effects of reservoir release when modeling the state transition, i.e., the state transition Eq.~\ref{eq:hid} becomes:
\begin{equation}
\begin{aligned}
\textbf{c}_i^t &= \text{tanh}(\textbf{gf}_i^t\otimes \textbf{c}_i^{t-1}+\textbf{gi}_i^t\otimes\bar{\textbf{c}}_i^t+\textbf{gs}_i^t\otimes \textbf{q}^{t-1}_i).\\
\end{aligned}
\label{eq:pretrain-hid}
\end{equation}

The model is pre-trained by minimizing the supervised loss function (Eq.~\ref{eq:std_loss}) using simulated temperature as training labels and the modified state transition equation (Eq.~\ref{eq:pretrain-hid}). After the pre-training, it can be fine-tuned using observations and water release data, by minimizing the loss defined in Eq.~\ref{eq:full_loss} and the original state equation (Eq.~\ref{eq:hid}).

\section{Experiments}
We evaluate our proposed method using data collected from the Delaware River Basin. In this section, we first introduce the dataset used for evaluation. Then we compare our method against multiple baselines in predicting water temperature with and without the information about reservoir releases. We further study the result of state adjustment mechanism using different update periods and different amounts of observation data. Finally, we validate the effectiveness of pre-training strategy when the model has limited training data.

\subsection{Dataset and Baselines}
\label{sec:dataset}
The dataset used in our evaluation is from U.S. Geological Survey's National Water Information System~\cite{us2016national} and the Water Quality Portal~\cite{read2017water}, the largest standardized water quality dataset for inland and coastal waterbodies~\cite{read2017water}. 
Observations at a specific latitude and longitude were matched to river segments that vary in length from 48 to 23,120 meters. The river segments were defined by the national geospatial fabric used for the National Hydrologic Model as described by Regan et al.~\cite{regan2018description_backup}, and the river segments are split up to have roughly a one day water travel time. We match observations to river segments by snapping observations to the nearest stream segment within a tolerance of 250 meters. Observations farther than 5,000 meters along the river channel to the outlet of a segment were omitted from our dataset. Segments with multiple observation sites were aggregated to a single mean daily water temperature value.    
 
We study a subset of the Delaware River Basin with 56 river segments  at Lordville, New York. We select this subset because we have sufficient observations collected in this area, 
which provides the flexibility to evaluate the capacity of different methods under data-sparse conditions by using a subset of data for training and testing. 
In particular, we use input features at the daily scale from Jan 01, 1980, to Mar 31, 2020 (14,700 dates). The input features have 10 dimensions, which include daily average precipitation, daily average air temperature, date of the year, solar radiation, shade fraction, potential evapotranspiration and the geometric features of each segment (e.g., elevation, length, slope and width). Air temperature, precipitation, and solar radiation values were derived from the gridMet gridded dataset~\cite{gridMET}. Other input features (e.g., shade fraction, potential evapotranspiration) are difficult to  measure frequently, and we use values internally calculated by the physics-based PRMS-SNTemp model~\cite{theurer1984instream}. Water temperature observations were available for 29 segments but the temperature was observed only on certain dates. The number of temperature observations  available 
for the 29 observed segments ranges from 1 to 13,000 with a total of 76,163  
observations across all dates and segments. In addition, we also have the water release data for the Cannonsville and Pepacton Reservoirs, which are in our study area. The release data includes how much water is released under each type of release strategy (conservation-based release, direct water release, and water spill release) at a daily scale. We also have meta-features of these reservoirs, including dam height, dam length, depth, elevation, and area of catchment.

We implement our method using Tensorflow and GTX2080 GPU. We have released our implementation and dataset\footnote{https://doi.org/10.5066/P9AHPO0H}. The model is optimized with the ADAM optimizer~\cite{kingma2014adam} with the initial learning rate of 0.002. All the hidden variables and gating variables in HRGN have 20 dimensions. The hyper-parameter $\lambda$ is set as 0.5. 
We generate the adjacency matrix $\textbf{A}$ based on the stream distance between each pair of nodes. When measuring the distance between a pair of river segments, we use the stream distance between their outlets. 
We represent the distance between node $i$ and node $j$ as $\text{dist}(i,j)$. We  standardize the stream distance and then compute the adjacency level as $\textbf{A}_{ij}=1/(1+\text{exp}(\text{dist}(i,j)))$ for each edge $(i,j)$.

We compare model performance to multiple baselines, including standard fully connected artificial neural networks (ANN), recurrent neural networks (RNN), recurrent graph neural networks (RGrN)~\cite{jia2021physics}. We also implement multiple versions of the proposed method. First, we implement our HRGN model without state adjustment. Then we implement Kalman Filter~\cite{bishop2001introduction} on our HRGN model as a baseline for state adjustment (HRGN-KF). Finally, we implement the HRGN with the proposed state adjustment mechanism and we name this method as HRGN-adj. We use HRGN-adj-$k$ to represent our implementation with adjustment using the update period of length $k$, i.e., the model state is adjusted only after every $k$ time steps.

In the following tests, we use data from Jan 01, 1980, to Oct 31, 2006, for training and then measure the testing performance on data from Nov 01, 2006, to Mar 31, 2020. The training of HRGN-adj is conducted in the same period but in two stages. We first optimize the HRGN without data assimilation mechanism  using training data from the period of Jan 01, 1980, to Jun 01, 1993. Then we apply the data assimilation mechanism using data from Jun 02, 1993, to Oct 31, 2006, i.e., we use adjusted state to compute the state and predictions at  next time step. We optimize the HRGN with state adjustment in this period. For all the methods with state adjustment (e.g., HRGN-KF and HRGN-adj), the state adjustment is performed during the testing period. This is not a fair comparison to other baselines because state adjustment uses data in the testing period for changing the model state. However, this is often feasible in real scenarios because these approaches only use observations from a historical view, i.e., they use observations for adjustment only after predictions have been made.

\subsection{Predictive Performance}

\begin{table}[!t]
\newcommand{\tabincell}[2]{\begin{tabular}{@{}#1@{}}#2\end{tabular}}
\centering
\caption{Predictive RMSE  for temperature prediction before and after we intentionally hide data for the Cannonsville Reservoir. For each testing scenario, we report the overall RMSE for all the river segments and the RMSE for the Segment X. }
\begin{tabular}{l|cc|cc}
\hline
& \multicolumn{2}{c|}{\textit{Before hiding}}&\multicolumn{2}{c}{\textit{After hiding}}\\
\textbf{Method} & \textbf{Overall} & \textbf{Segment X}  &  \textbf{Overall} &  \textbf{Segment X}\\ \hline 
ANN & 2.16& 2.30& 2.16& 2.30\\
RNN & 1.92& 2.18& 1.92& 2.18\\
RGrN & 1.86 & 2.08& 1.86 & 2.08\\
HRGN   & 1.41 & 1.58& 1.80 & 2.09 \\ \hline 
HRGN-KF &1.25 & 1.35& 1.65&1.89\\
HRGN-adj& 0.75 & 0.64 & 0.76 & 0.66\\
HRGN-adj-7&  1.24 & 1.30& 1.26 & 1.34\\
\hline
\end{tabular}
\label{perf_temp}
\end{table}

Here we conduct two sets of experiment. First, we train and test each method using all the available  data. Then we intentionally hide the reservoir release data from the Cannonsville Reservoir (for both training and testing periods) and  evaluate all the methods again. When measuring the performance of each method, we compute the overall Root Mean Squared Error (RMSE) over all the river segments in the testing period and also the RMSE on the river segment (termed ``Segment X''), which is directly downstream from the Cannonsville Reservoir. This segment is most affected by the reservoir, and the predictions on this segment are therefore most likely to be affected if the model does not represent the water released from the Cannonsville Reservoir. 

\begin{figure*} [!h] 
\centering
\subfigure[]{ \label{fig:a}{}
\includegraphics[width=0.45\textwidth]{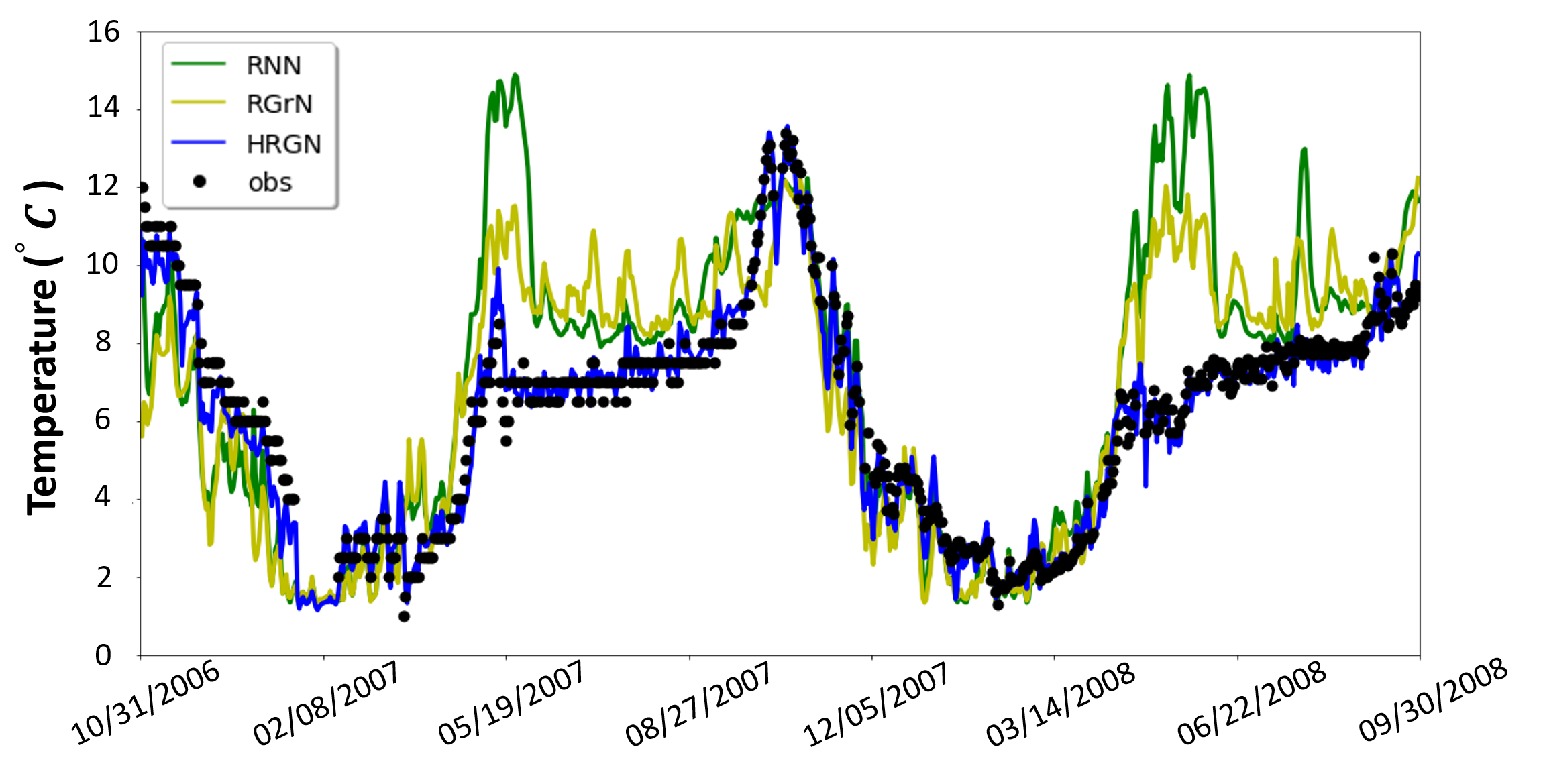}
}
\subfigure[]{ \label{fig:b}{}
\includegraphics[width=0.45\textwidth]{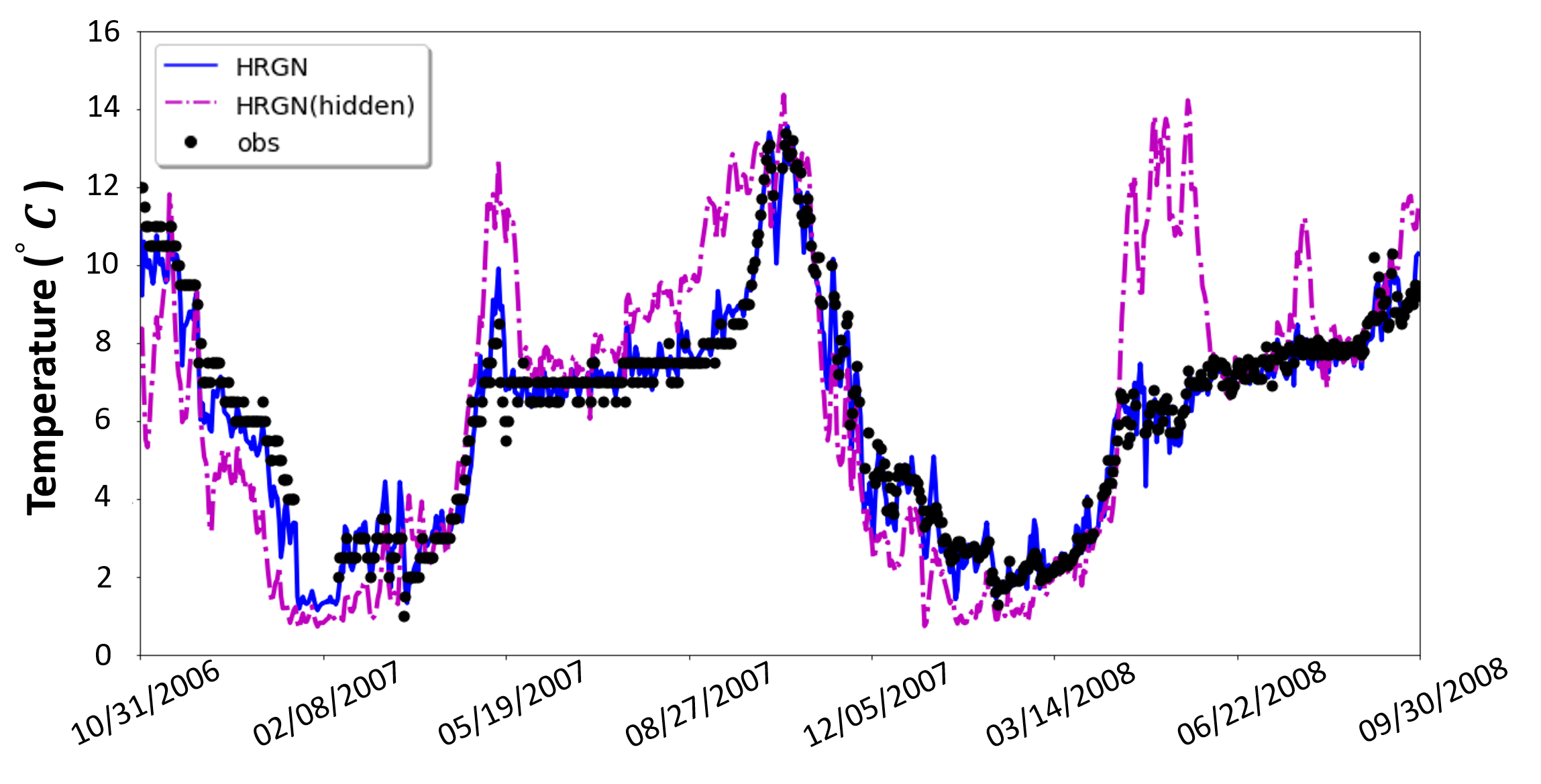}
}
\\
\subfigure[]{ \label{fig:b}{}
\includegraphics[width=0.45\textwidth]{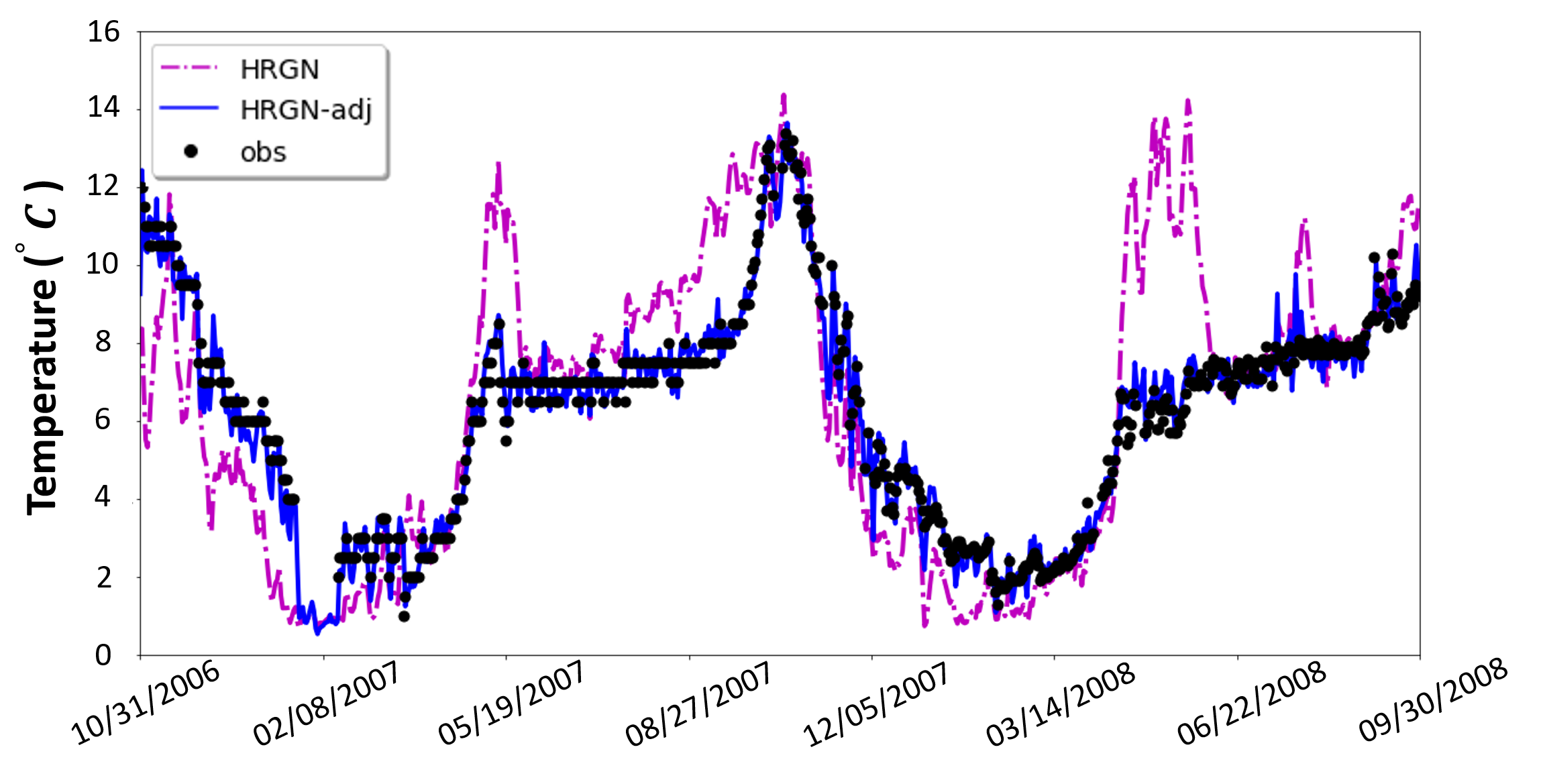}
}
\subfigure[]{ \label{fig:b}{}
\includegraphics[width=0.45\textwidth]{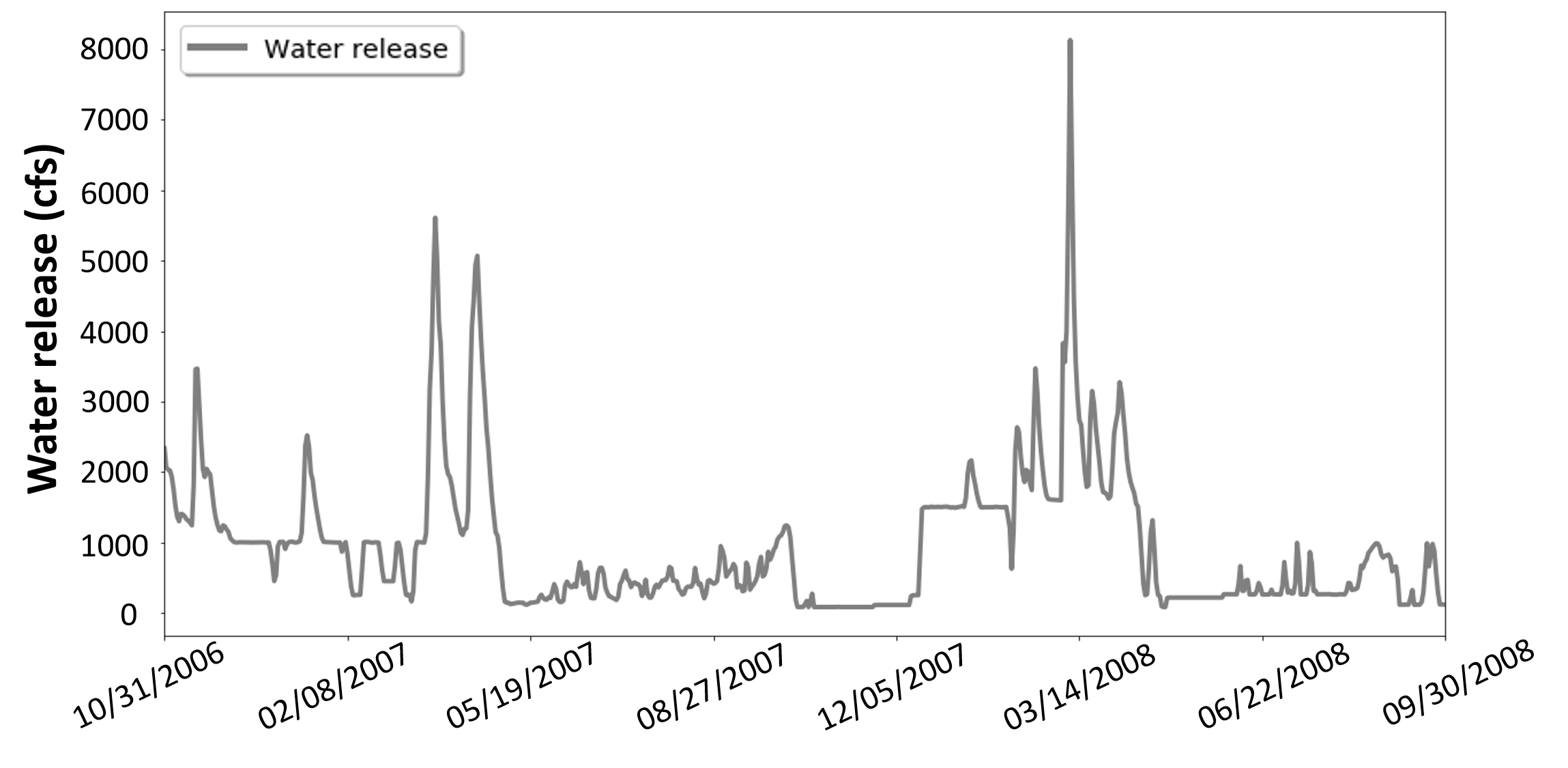}
}
\caption{(a) Predictions made by RNN, RGrN, and HRGN when the reservoir release data are available. (b) Predictions made by HRGN before  and after  we hide the reservoir release data. (c) Predictions made by HRGN and HRGN-adj when the Cannonsville Reservoir release data are missing. (d) The amount of water release (aggregated over different release types) over the same period with (a)-(c).}
\label{fig:prediction}
\end{figure*}

We report the performance of different models in these two tests in Table~\ref{perf_temp}. The ANN, RNN, and RGrN model do not take into account reservoir release so they have the same performance before and after we hide reservoir release data. When we have all the reservoir data available, we notice that our proposed HRGN outperforms other baselines (ANN, RNN, and RGrN) when no adjustment is performed. The improvement from ANN to RNN and from RNN to RGrN shows that the effectiveness of leveraging temporal dependencies and spatial dependencies in simulating water temperature dynamics. Moreover, the comparison between RGrN and HRGN confirms that incorporating water release data helps improve the predictive performance. In Fig.~\ref{fig:prediction} (a), we show the predictions made by RNN, RGrN, and HRGN in Segment X using all the reservoir data. We can clearly see that RNN and RGrN over-estimate water temperature at the beginning of summer period while HRGN predictions match observations much better in these periods. This is because the amount of water release is much higher at the beginning of summer (Fig.~\ref{fig:prediction} (d)) and the baseline models cannot capture how such water release cools downstream segments.

After we hide the reservoir release data, we can see that HRGN results in larger errors (by comparing HRGN in the column 1-2 and column 3-4 in Table~\ref{perf_temp}). Fig.~\ref{fig:prediction} shows the predictions made by HRGN before and after we hide the reservoir release data. The HRGN over-predicts water temperature when it is not aware of the release information. 

In Table~\ref{perf_temp}, we also see that the methods using state adjustment mechanisms (HRGN-KF, HRGN-adj, and HRGN-adj-7) perform much better than methods without using the adjustment, especially after we hide reservoir release data. This is also confirmed by the predictions made by HRGN and HRGN-adj as shown in Fig.~\ref{fig:prediction} (c). 
The performance of the state adjustment approach shows the value of using observations (when available) to correct model bias for improved short-term predictions of temperature. It can also be seen that HRGN-adj performs better than HRGN-KF, which confirms the effectiveness of our proposed invertible structure. Furthermore, we see that  HRGN-adj performs worse when using a longer update period (e.g., HRGN-adj-7). This is because the model state is updated less frequently compared to HRGN-adj. More details on the effect of the update period is given in Section~\ref{sec:adjustdata}.

\subsection{State Adjustment under Different Scenarios}
\label{sec:adjustdata}

\begin{figure} [!h] 
\centering
\subfigure[]{ \label{fig:a}{}
\includegraphics[width=0.23\textwidth]{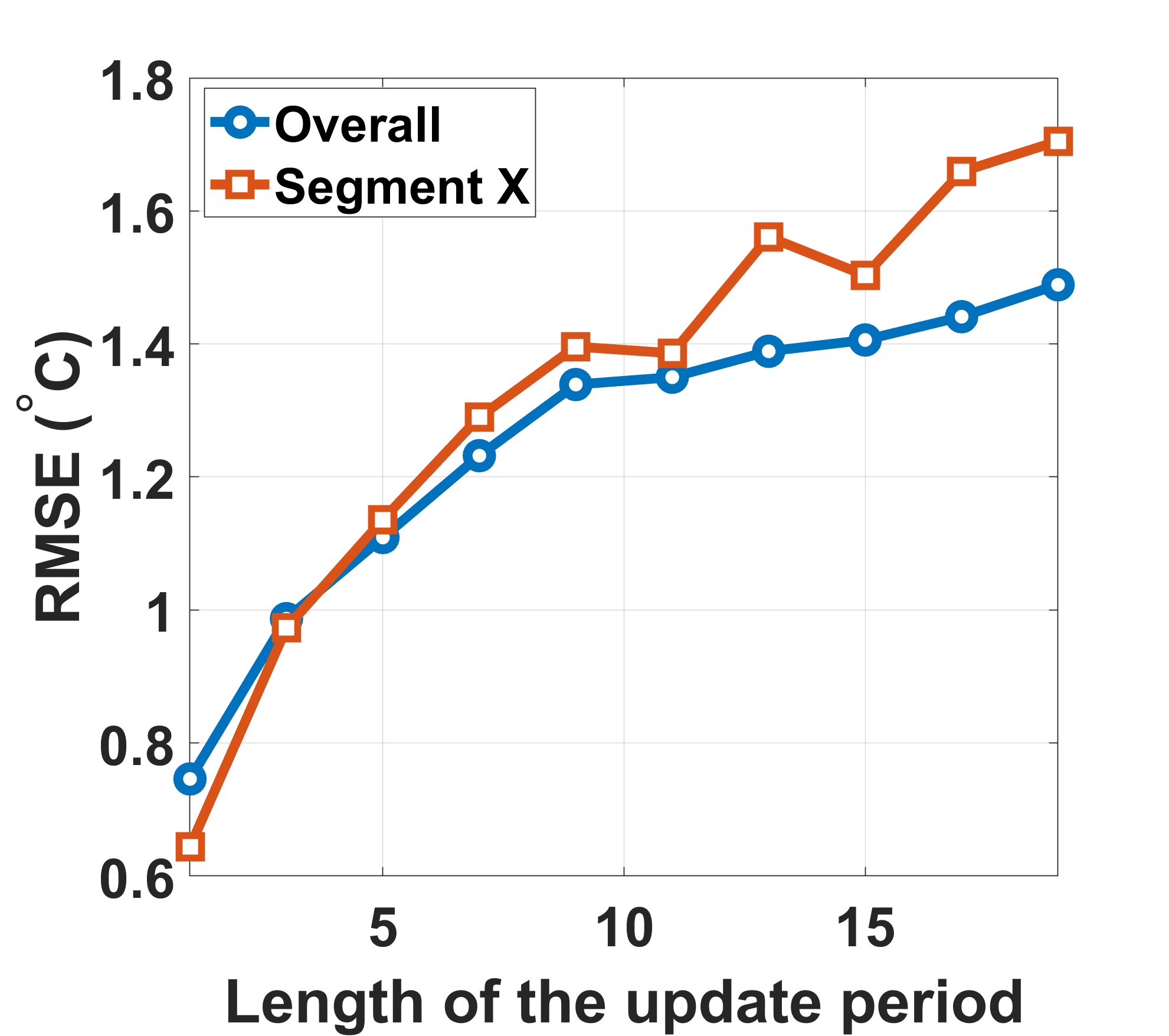}
}\hspace{-0.25in}
\subfigure[]{ \label{fig:b}{}
\includegraphics[width=0.23\textwidth]{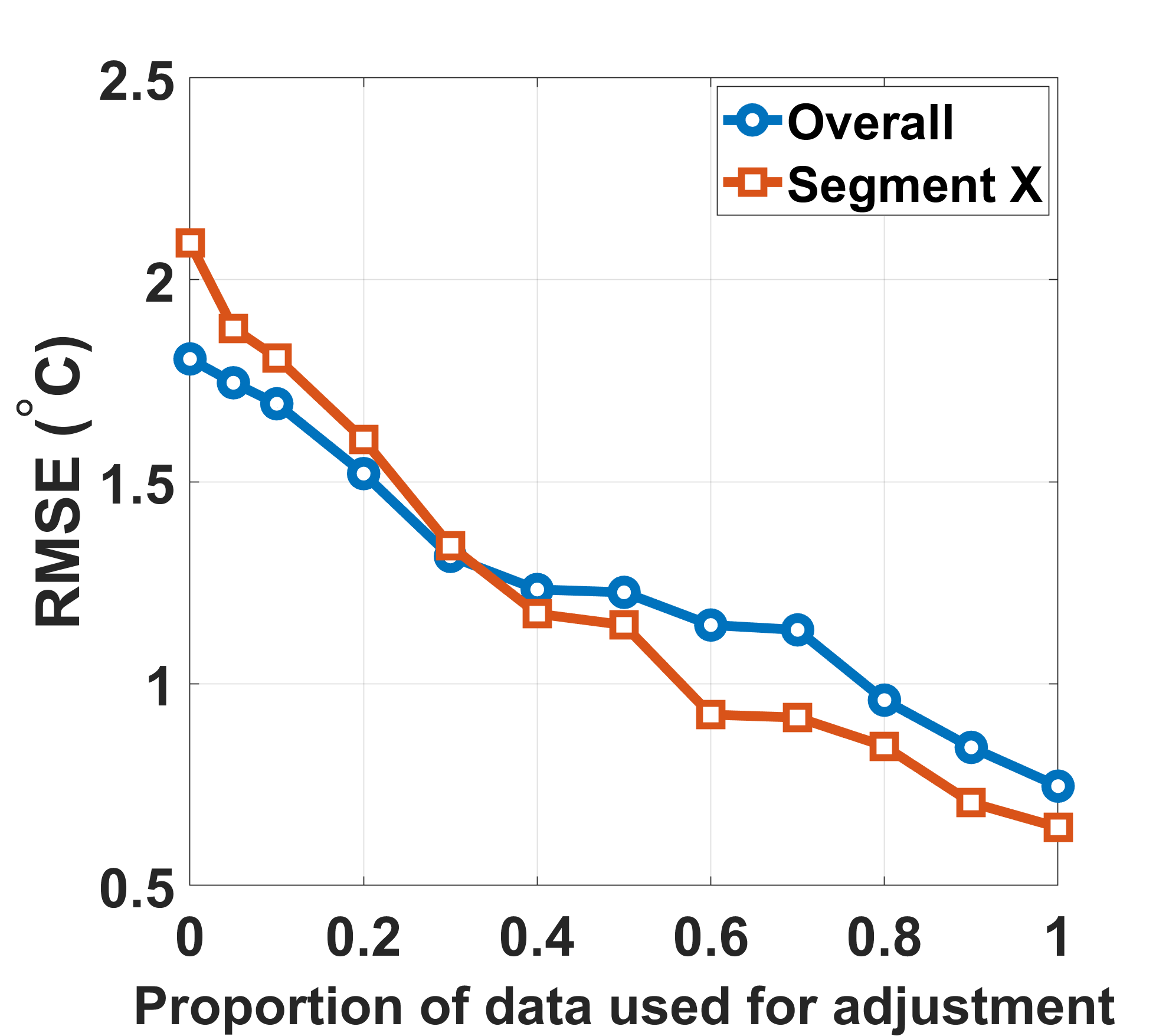}
}%
\vspace{-.1in}
\caption{(a) Change of predictive error when we use different update periods. (b) Change of predictive error when we use different amounts of observations for state adjustment.} 
\label{fig:adjustdata}
\vspace{-.2in}
\end{figure}

In Fig.~\ref{fig:adjustdata}, we show the change of performance as we change the update period (Fig.~\ref{fig:adjustdata} (a)) and the amount of data used for adjustment (Fig.~\ref{fig:adjustdata} (b)). When we reduce the data in Fig.~\ref{fig:adjustdata} (b), we randomly select a proportion of data from the period Nov 01, 2006, to Mar 31, 2020, for adjusting the model state. We still use all the observations in the testing period for evaluating the predictive RMSE.

\begin{figure*} [!h] 
\centering
\subfigure[]{ \label{fig:a}{}
\includegraphics[width=0.45\textwidth]{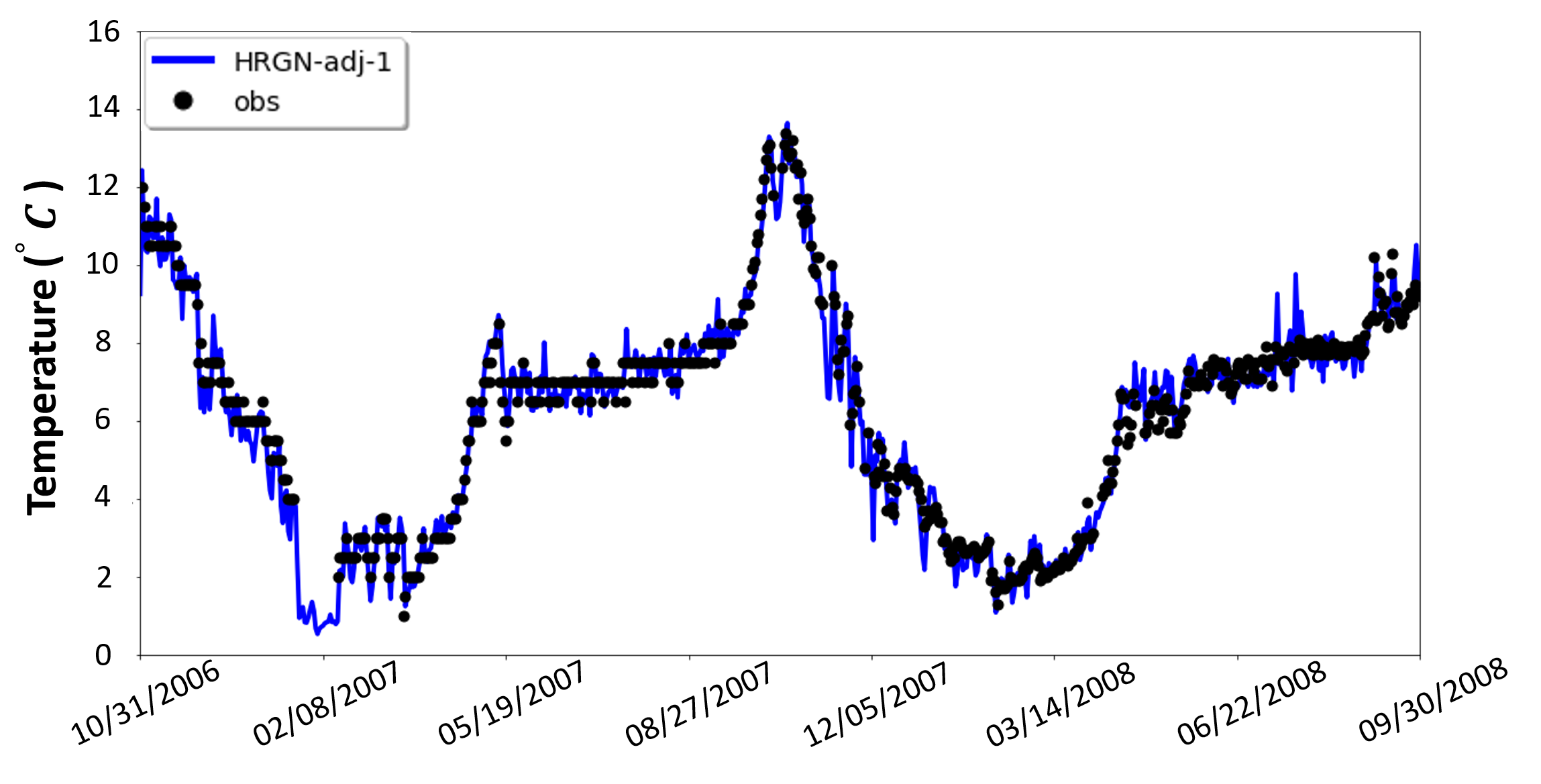}
} 
\subfigure[]{ \label{fig:b}{}
\includegraphics[width=0.45\textwidth]{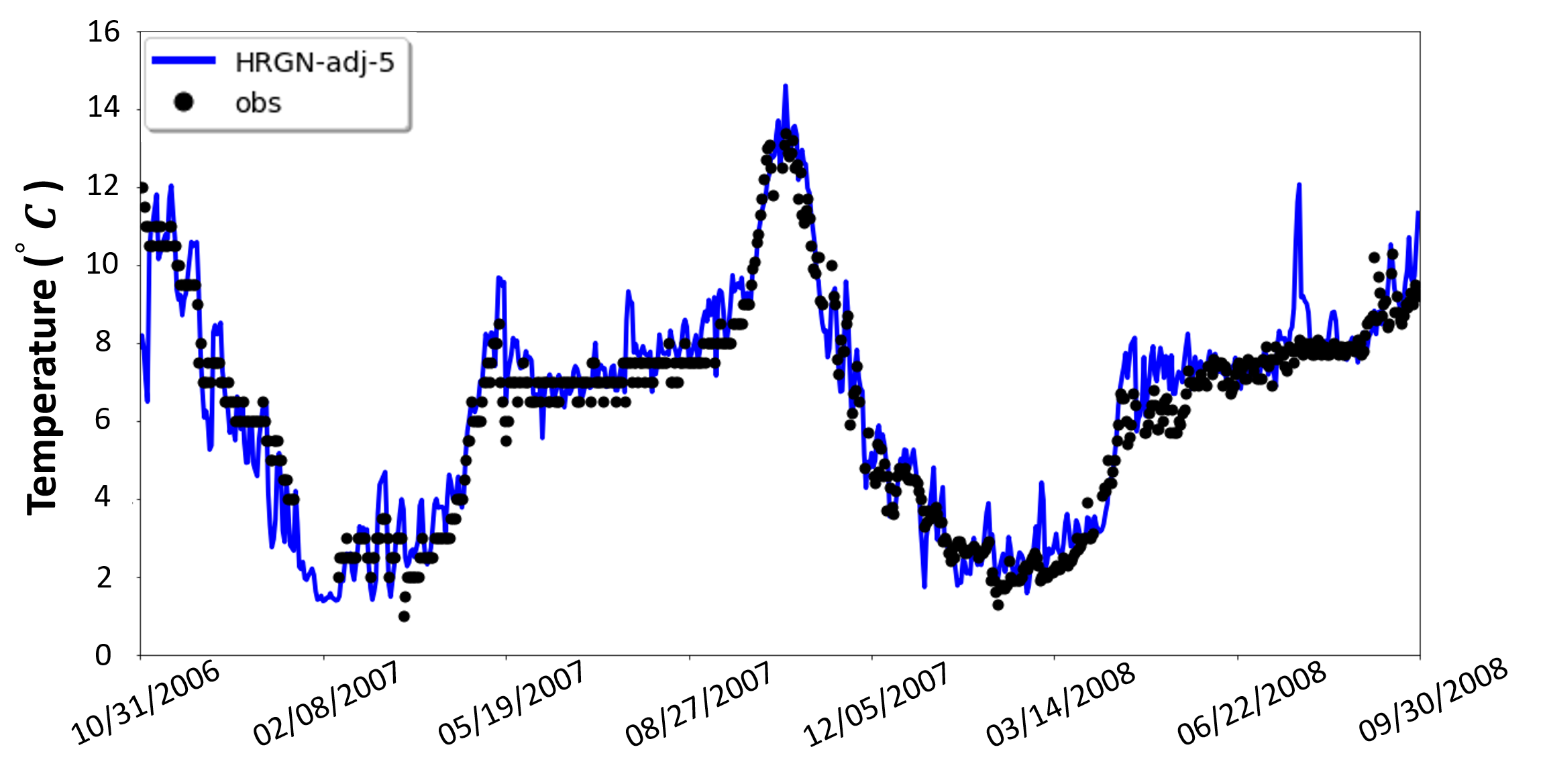}
}\\\vspace{-.1in}
\subfigure[]{ \label{fig:b}{}
\includegraphics[width=0.45\textwidth]{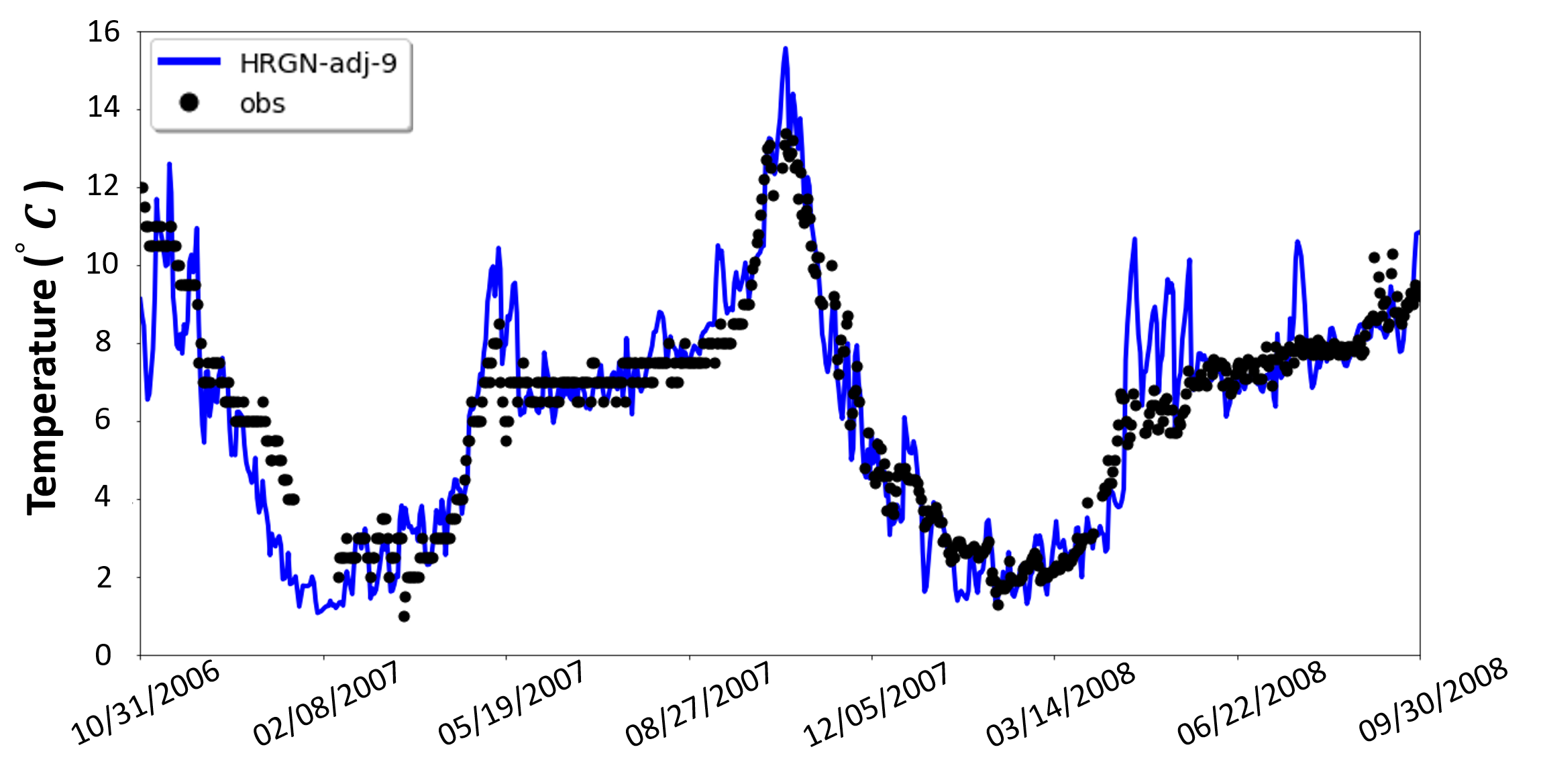}
}
\subfigure[]{ \label{fig:b}{}
\includegraphics[width=0.45\textwidth]{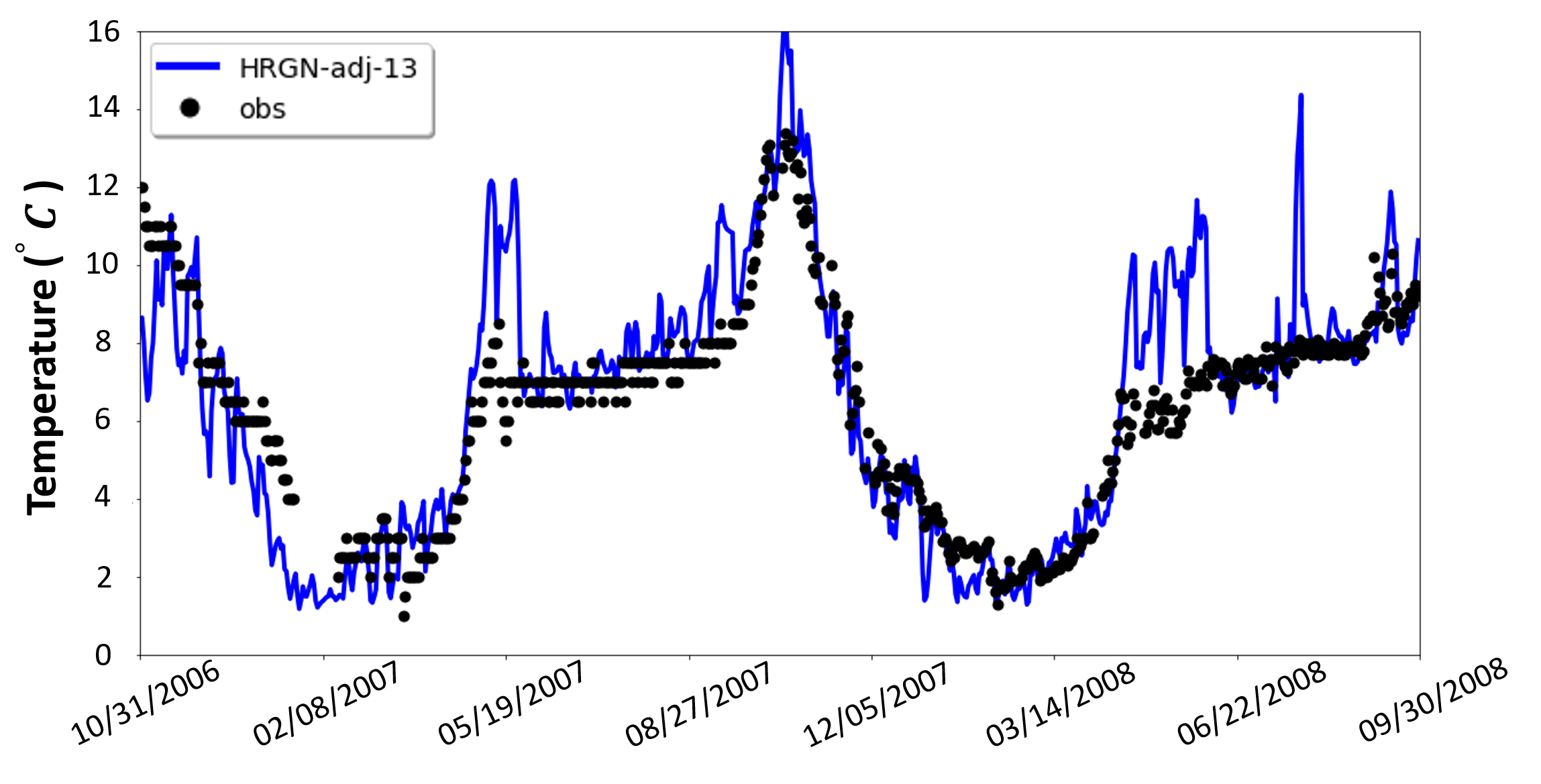} }
\vspace{-.1in}
\caption{Predictions made by HRGN-adj using (a) 1-day update period, (b) 5-day update period, (c) 9-day update period, and (d) 13-day update period, when the Cannonsville Reservoir release data are missing.}
\label{fig:hrgn-window}
\vspace{-.2in}
\end{figure*}

We can see that the adjustment using longer update periods leads to larger predictive errors. We visualize predictions made by HRGN-adj using different update periods in Fig.~\ref{fig:hrgn-window} (with Cannonsville Reservoir release data hidden). It can be seen that HRGN-adj-$k$ with a large $k$ value has a larger bias in predictions, especially at the beginning of summer periods. We can also see that there is a clear delay of model adjustment as we increase the update period.

\begin{figure} [!h] 
\centering
\subfigure[]{ \label{fig:a}{}
\includegraphics[width=0.45\textwidth]{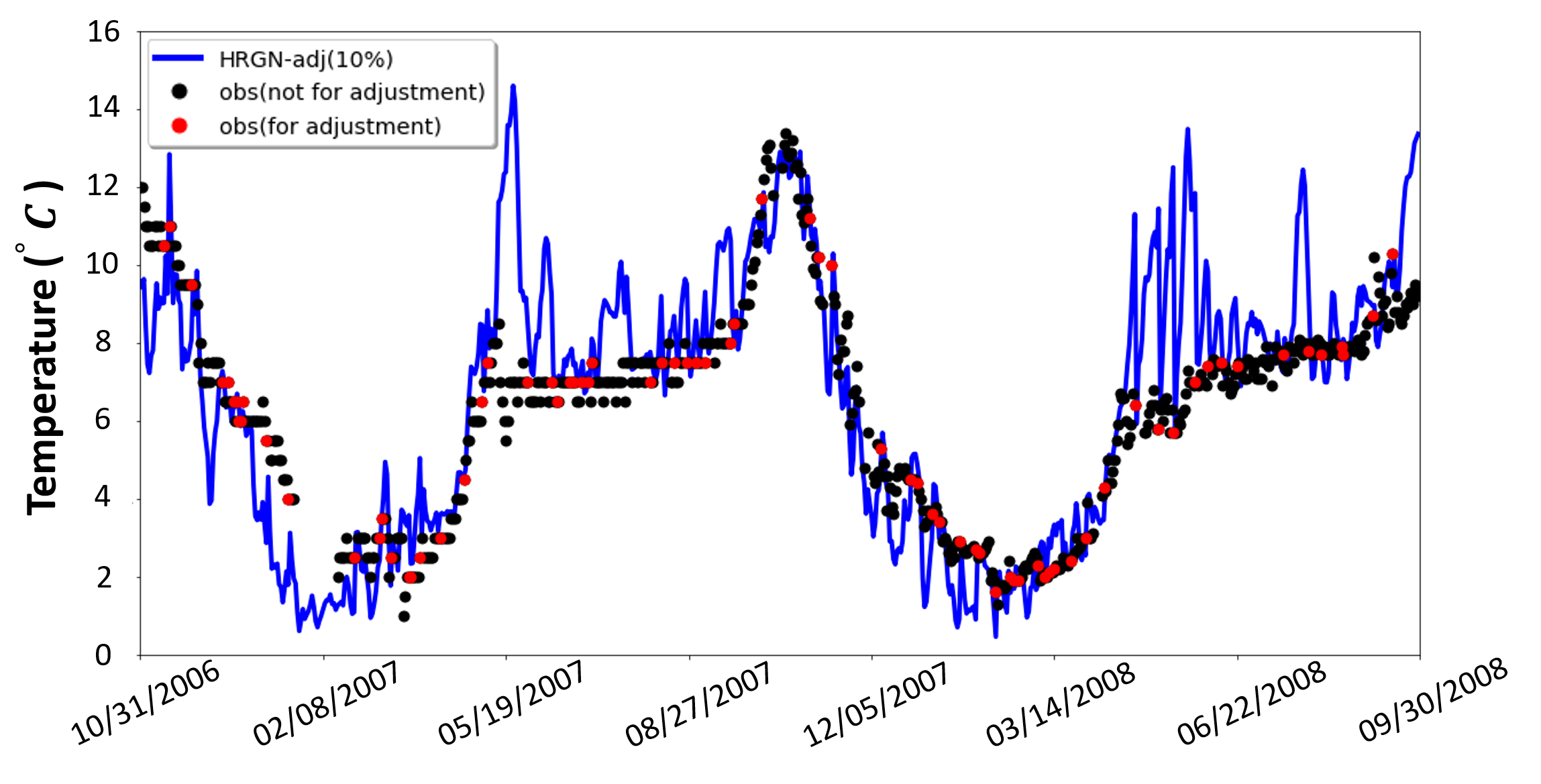}
} \\
\vspace{-.1in}
\subfigure[]{ \label{fig:b}{}
\includegraphics[width=0.45\textwidth]{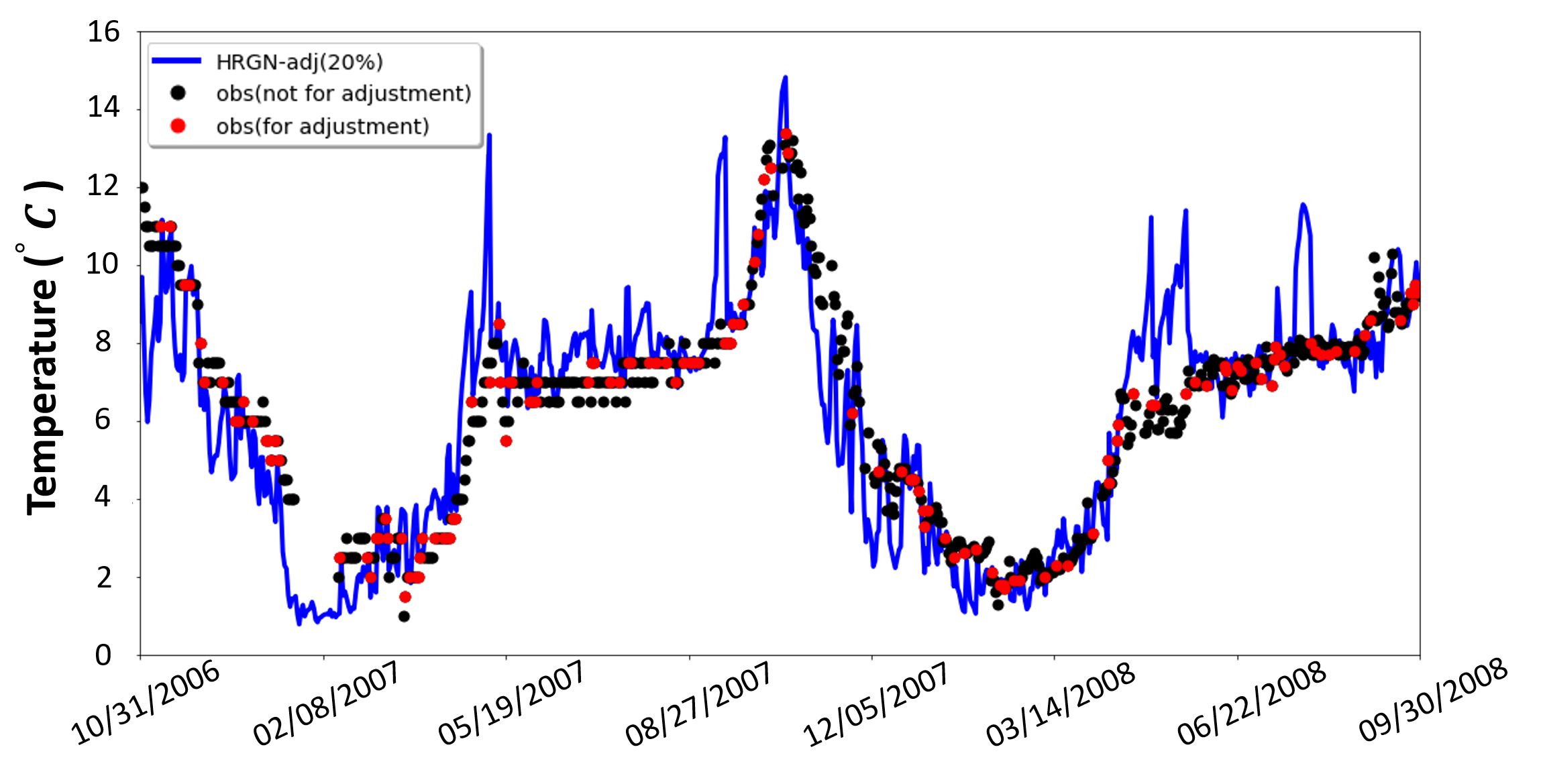}
}\\
\caption{Predictions made by HRGN-adj using (a) 10\% and (b) 20\% observations from Jun 02, 1993, to Oct 31, 2006, when the Cannonsville Reservoir release data is missing.}
\label{fig:hrgn-data}
\vspace{-.2in}
\end{figure}

Similarly, Fig.~\ref{fig:adjustdata} shows that HRGN-adj has worse performance as we reduce the amount of observations used for adjustment. We show the predictions made by HRGN-adj using 10\% and 20\% data in Fig.~\ref{fig:hrgn-data}(a) and Fig.~\ref{fig:hrgn-data}(b), respectively. It can be seen that the model cannot adjust its state frequently because observations (marked in red color) are often not available. 
However, once the model has access to an observed data point, the adjustment alone can lead to much better predictions at the next time step.

\subsection{Pre-training for Handling Sparse Training Observations}

Here we validate the effectiveness of the pre-training method when the model is trained with fewer labeled samples. In particular, we randomly select a proportion of observations from the training period (Jan 01, 1980, to Oct 31, 2006) to withhold from model training and then test the performance HRGN and HRGN-adj. Moreover, we use simulated water temperature data produced by the PRMS-SNTemp model~\cite{theurer1984instream,markstrom2015prms} during the entire study period to pre-train HRGN and HRGN-adj. We use HRGN$^\text{ptr}$ (and HRGN-adj$^\text{ptr}$) to represent the HRGN model (and HRGN-adj model) pre-trained using simulations and then fine-tuned using observations. In Table~\ref{perf_sparsetrain}, we can see that all the methods get worse performance as we reduce the amount of training data. We also notice that {HRGN-adj} can get even worse performance than HRGN when we use an extremely  small amount of data (0.1\%). This is because the model cannot learn a good state transition relation from one time to the next time step and thus the state adjustment is not helpful for improving future predictions.

\begin{figure} [!h] 
\centering
\includegraphics[width=0.5\textwidth]{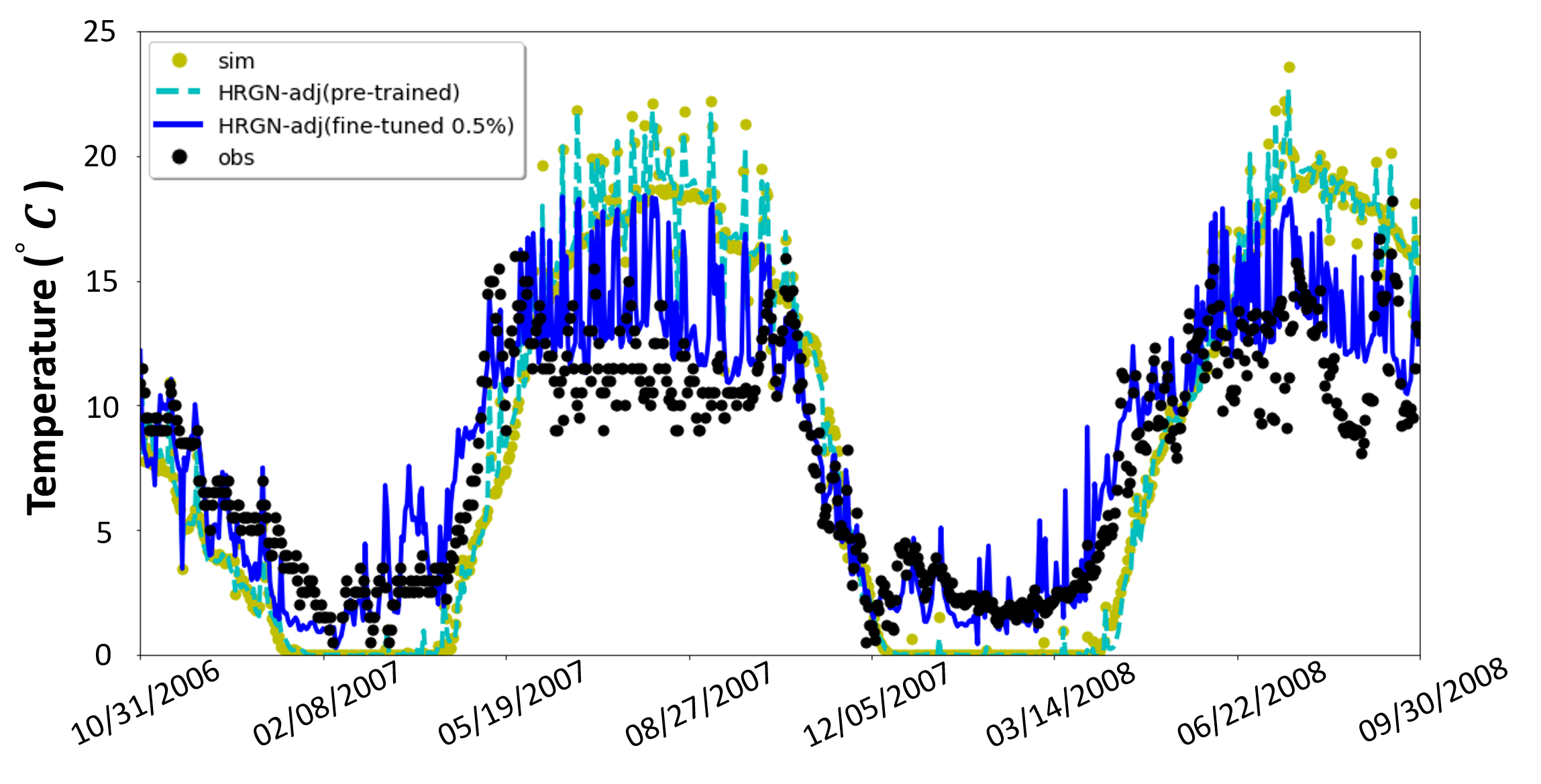}
\caption{Predictions made by (a) HRGN-adj pre-trained using PRMS-SNTemp simulations and (b) HRGN-adj pre-trained and then fine-tuned using 0.5\% training observations. We also show the PRMS-SNTemp simulations in yellow. }
\label{fig:hrgn-pretrain}
\vspace{-.2in}
\end{figure}

With the pre-training strategy, we can clearly see that HRGN and HRGN-adj perform better given limited training data. We also report the RMSE of the simulation data used for pre-training in the first row of Table~\ref{perf_sparsetrain}. Although the physical simulations have a large bias, they are still helpful when used for pre-training the HRGN model. In Fig.~\ref{fig:hrgn-pretrain}, we show predictions  made by HRGN-adj after being pre-trained (without fine-tuning) and predictions made by the same model after being fine-tuned with 0.5\% observations on a specific river segment. It can be seen that the pre-trained model matches with simulations very well. Although the simulations have a large gap to true observations, they can still represent some general physical relations and patterns that are used to build the physics-based model. The HRGN-adj model pre-trained using simulations has a higher chance at capturing such patterns (e.g., seasonal cycles) and thus require much less data for fine-tuning. We can see that the model fine-tuned with just 0.5\% data can match true observations better. 

This figure also highlights the effect that the HGRN has in representing the effect of reservoirs. The PRMS-SNTemp model does not represent the reservoirs, therefore the simulated water tempeartures follow a sinosouidal-like pattern, with peak tempeartures in the summer months. The observations, however, show the effect of the reservoir - the observed temperatures level off in the summertime. With just 0.5\% of the observations, the HGRN is able to start to follow the observed pattern of lower summer temperatures that occurs due to the reservoir releases. It also starts to pick up that the water is warmer in the winter due to winter reservoir releases.

\begin{table}[!t]
\footnotesize
\newcommand{\tabincell}[2]{\begin{tabular}{@{}#1@{}}#2\end{tabular}}
\centering
\caption{Prediction RMSE  for temperature modeling using different amounts of training observations. Rows in gray color represent methods that are first pre-trained using simulation data and then fine-tuned using training observations. We also report RMSE for simulations produced by the physics-based PRMS-SNTemp model (first row). The PRMS-SNTemp model is uncalibrated so the performance does not change with different training data.}
\begin{tabular}{l|ccccccc}
\hline
\textbf{Method} & 0.1\%&0.5\% & 2\% & 10\%&  20\% & 50\%& 100\% \\ \hline 
PRMS-SNTemp & 4.05 &4.05&4.05&4.05&4.05&4.05&4.05\\ 
HRGN & 3.61& 3.05&2.63& 2.43& 1.87&1.86&1.86\\
\rowcolor[gray]{0.95}
HRGN$^\text{ptr}$ & 2.90 & 2.82& 2.41& 1.88 & 1.80& 1.79& 1.78\\
\hline
HRGN-adj& 4.03 &3.01& 2.80& 1.51& 1.44& 1.10& 0.75 \\
\rowcolor[gray]{0.95}
HRGN-adj$^\text{ptr}$& 3.35 &2.46 & 2.47&1.41 & 1.47& 0.90 & 0.74\\
\hline
\end{tabular}
\label{perf_sparsetrain}
\end{table}

\section{Conclusions}

In this paper, we develop an HRGN model to represent interacting processes amongst river segments and reservoirs. When reservoir release data are not available, we also introduce a data assimilation mechanism to adjust the model state using observations. Finally, we use a pre-training strategy to ensure good model quality that is critical for the assimilation mechanism. We have several observations from our experiments for the Delaware River Basin: (1) The HRGN model outperforms existing methods in predicting stream temperature when reservoir data are available. (2) The data assimilation mechanism using fewer observations or a longer update period will increase the prediction error. (3) Despite reduced performance with fewer observations or a longer update period, the data assimilation mechanism can still improve the performance, and has an especially large effect when the reservoir release information is missing. (4) The data assimilation mechanism leads to worse performance when the model is poorly trained (with limited data) but the pre-training strategy can effectively alleviate this issue. 

While our method is developed in the context of modeling stream-reservoir networks, the HRGN model and the data assimilation mechanism are applicable in a broad range of scientific applications. For example, the HRGN model can be potentially used for modeling interactions amongst different types of devices in power systems. The state adjustment approach can also be used to improve the existing data assimilation methods used in climate science and Earth science, which are commonly expensive to implement. 
It would be beneficial to extend the data assimilation mechanism to adjust model states for nodes that are connected to the observed node in the graph, which is critical when observations are sparse over space.  


\section{acknowledgements}


This work was support by the USGS Award G21AC10207 and Pitt Momentum Award. This research was supported in part by the University of Pittsburgh Center for Research Computing through the resources provided. 
Any use of trade, firm, or product names is for descriptive purposes only and does not imply endorsement by the U.S. Government.

\bibliographystyle{IEEEtran}
\bibliography{IEEEabrv}

\begin{thebibliography}{10}
\providecommand{\url}[1]{#1}
\csname url@samestyle\endcsname
\providecommand{\newblock}{\relax}
\providecommand{\bibinfo}[2]{#2}
\providecommand{\BIBentrySTDinterwordspacing}{\spaceskip=0pt\relax}
\providecommand{\BIBentryALTinterwordstretchfactor}{4}
\providecommand{\BIBentryALTinterwordspacing}{\spaceskip=\fontdimen2\font plus
\BIBentryALTinterwordstretchfactor\fontdimen3\font minus
  \fontdimen4\font\relax}
\providecommand{\BIBforeignlanguage}[2]{{%
\expandafter\ifx\csname l@#1\endcsname\relax
\typeout{** WARNING: IEEEtran.bst: No hyphenation pattern has been}%
\typeout{** loaded for the language `#1'. Using the pattern for}%
\typeout{** the default language instead.}%
\else
\language=\csname l@#1\endcsname
\fi
#2}}
\providecommand{\BIBdecl}{\relax}
\BIBdecl

\bibitem{brett1971energetic}
J.~R. Brett, ``Energetic responses of salmon to temperature. a study of some
  thermal relations in the physiology and freshwater ecology of sockeye salmon
  (oncorhynchus nerka),'' \emph{American zoologist}, 1971.

\bibitem{theurer1984instream}
F.~Theurer, K.~Voos, and W.~Miller, ``Instream water temperature model.
  instream flow information paper 16. us fish wildl serv,'' \emph{Div. Biol.
  Serv., Tech. Rep. FWS OBS}, vol.~84, no.~15, pp. 11--42, 1984.

\bibitem{markstrom2015prms}
S.~L. Markstrom, R.~S. Regan, L.~E. Hay, R.~J. Viger, R.~M. Webb, R.~A. Payn,
  and J.~H. LaFontaine, ``Prms-iv, the precipitation-runoff modeling system,
  version 4,'' \emph{US Geological Survey Techniques and Methods}, no. 6-B7,
  2015.

\bibitem{moshe2020hydronets}
Z.~Moshe, A.~Metzger, G.~Elidan, F.~Kratzert, S.~Nevo, and R.~El-Yaniv,
  ``Hydronets: Leveraging river structure for hydrologic modeling,''
  \emph{arXiv preprint arXiv:2007.00595}, 2020.

\bibitem{jia2021physics}
X.~Jia, J.~Zwart, J.~Sadler, A.~Appling, S.~Oliver, S.~Markstrom, J.~Willard,
  S.~Xu, M.~Steinbach, J.~Read \emph{et~al.}, ``Physics-guided recurrent graph
  model for predicting flow and temperature in river networks,'' in
  \emph{SDM}.\hskip 1em plus 0.5em minus 0.4em\relax SIAM, 2021, pp. 612--620.

\bibitem{kipf2016semi}
T.~N. Kipf and M.~Welling, ``Semi-supervised classification with graph
  convolutional networks,'' \emph{arXiv preprint arXiv:1609.02907}, 2016.

\bibitem{hamilton2017inductive}
W.~L. Hamilton, R.~Ying, and J.~Leskovec, ``Inductive representation learning
  on large graphs,'' \emph{arXiv preprint arXiv:1706.02216}, 2017.

\bibitem{qi2019hybrid}
Y.~Qi, Q.~Li, H.~Karimian, and D.~Liu, ``A hybrid model for spatiotemporal
  forecasting of pm2. 5 based on graph convolutional neural network and long
  short-term memory,'' \emph{Science of the Total Environment}, 2019.

\bibitem{xie2018crystal}
T.~Xie and J.~C. Grossman, ``Crystal graph convolutional neural networks for an
  accurate and interpretable prediction of material properties,''
  \emph{Physical review letters}, vol. 120, no.~14, p. 145301, 2018.

\bibitem{zhu2020understanding}
D.~Zhu \emph{et~al.}, ``Understanding place characteristics in geographic
  contexts through graph convolutional neural networks,'' \emph{Annals of the
  American Association of Geographers}, 2020.

\bibitem{shi2016survey}
C.~Shi, Y.~Li, J.~Zhang, Y.~Sun, and S.~Y. Philip, ``A survey of heterogeneous
  information network analysis,'' \emph{IEEE Transactions on Knowledge and Data
  Engineering}, vol.~29, no.~1, pp. 17--37, 2016.

\bibitem{zhang2019heterogeneous}
C.~Zhang, D.~Song, C.~Huang, A.~Swami, and N.~V. Chawla, ``Heterogeneous graph
  neural network,'' in \emph{SIGKDD}, 2019.

\bibitem{wang2019heterogeneous}
X.~Wang, H.~Ji, C.~Shi, B.~Wang, Y.~Ye, P.~Cui, and P.~S. Yu, ``Heterogeneous
  graph attention network,'' in \emph{WWW}, 2019.

\bibitem{zhu2020hgcn}
Z.~Zhu, X.~Fan, X.~Chu, and J.~Bi, ``Hgcn: A heterogeneous graph convolutional
  network-based deep learning model toward collective classification,'' in
  \emph{SIGKDD}, 2020.

\bibitem{evensen2009data}
G.~Evensen, \emph{Data assimilation: the ensemble Kalman filter}.\hskip 1em
  plus 0.5em minus 0.4em\relax Springer Science \& Business Media, 2009.

\bibitem{fang2020near}
K.~Fang and C.~Shen, ``Near-real-time forecast of satellite-based soil moisture
  using long short-term memory with an adaptive data integration kernel,''
  \emph{Journal of Hydrometeorology}, vol.~21, no.~3, pp. 399--413, 2020.

\bibitem{brajard2020combining}
J.~Brajard, A.~Carrassi, M.~Bocquet, and L.~Bertino, ``Combining data
  assimilation and machine learning to emulate a dynamical model from sparse
  and noisy observations: a case study with the lorenz 96 model,''
  \emph{Journal of Computational Science}, vol.~44, p. 101171, 2020.

\bibitem{tan2018survey}
C.~Tan, F.~Sun, T.~Kong, W.~Zhang, C.~Yang, and C.~Liu, ``A survey on deep
  transfer learning,'' in \emph{International conference on artificial neural
  networks}.\hskip 1em plus 0.5em minus 0.4em\relax Springer, 2018, pp.
  270--279.

\bibitem{read2019process}
J.~S. Read \emph{et~al.}, ``Process-guided deep learning predictions of lake
  water temperature,'' \emph{Water Resources Research}, 2019.

\bibitem{jia2021physics_tds}
X.~Jia, J.~Willard, A.~Karpatne, J.~S. Read, J.~A. Zwart, M.~Steinbach, and
  V.~Kumar, ``Physics-guided machine learning for scientific discovery: An
  application in simulating lake temperature profiles,'' \emph{ACM/IMS
  Transactions on Data Science}, vol.~2, no.~3, pp. 1--26, 2021.

\bibitem{dinh2016density}
L.~Dinh, J.~Sohl-Dickstein, and S.~Bengio, ``Density estimation using real
  nvp,'' \emph{arXiv preprint arXiv:1605.08803}, 2016.

\bibitem{ardizzone2019guided}
L.~Ardizzone, C.~L{\"u}th, J.~Kruse, C.~Rother, and U.~K{\"o}the, ``Guided
  image generation with conditional invertible neural networks,'' \emph{arXiv
  preprint arXiv:1907.02392}, 2019.

\bibitem{us2016national}
U.~$\hspace{-.02in}$S. Geological{$\,\,$}Survey, ``National water information
  system data available on the world wide web (usgs water data for the nation),
  https://doi.org/10.5066/f7p55kjn,'' 2021.

\bibitem{read2017water}
E.~K. Read, L.~Carr, L.~De~Cicco, H.~A. Dugan, P.~C. Hanson, J.~A. Hart,
  J.~Kreft, J.~S. Read, and L.~A. Winslow, ``Water quality data for
  national-scale aquatic research: The water quality portal,'' \emph{Water
  Resources Research}, vol.~53, no.~2, pp. 1735--1745, 2017.

\bibitem{regan2018description_backup}
R.~S. Regan, S.~L. Markstrom, L.~E. Hay, R.~J. Viger, P.~A. Norton, J.~M.
  Driscoll, and J.~H. LaFontaine, ``Description of the national hydrologic
  model for use with the precipitation-runoff modeling system (prms),'' U.S.
  Geological Survey, Tech. Rep., 2018.

\bibitem{gridMET}
``{gridMET - Climatology Lab},''
  \url{http://www.climatologylab.org/gridmet.html}, accessed: 2001-05-01.

\bibitem{kingma2014adam}
D.~P. Kingma and J.~Ba, ``Adam: A method for stochastic optimization,''
  \emph{arXiv preprint arXiv:1412.6980}, 2014.

\bibitem{bishop2001introduction}
G.~Bishop, G.~Welch \emph{et~al.}, ``An introduction to the kalman filter,''
  \emph{Proc of SIGGRAPH, Course}, vol.~8, no. 27599-23175, p.~41, 2001.

\end{thebibliography}

\end{document}